\documentclass[sigconf]{acmart}

\AtBeginDocument{%
  }

\copyrightyear{2026}
\acmYear{2026}
\setcopyright{cc}
\setcctype{by}
\acmConference[CIKM '26]{Proceedings of the 35th ACM International Conference on Information and Knowledge Management}{November 07--11, 2026}{Rome, Italy}
\acmBooktitle{Proceedings of the 35th ACM International Conference on Information and Knowledge Management (CIKM '26), November 07--11, 2026, Rome, Italy}
\acmDOI{10.1145/3799682.3840622}
\acmISBN{979-8-4007-2539-5/2026/11}

\usepackage{algorithm}
\usepackage{algorithmic}

\acmSubmissionID{cfp0774}

\begin{document}

\title[Degree-Mass Message Passing for Betweenness Ranking]{Degree-Mass Message Passing for Betweenness Ranking in Directed and Undirected Networks}

\author{Justin Dachille}
\authornote{Both authors contributed equally to this research.}
\orcid{0009-0005-5390-7233}
\affiliation{%
  \institution{King's College London}
  \city{London}
  \country{United Kingdom}
}
\email{justin.dachille@kcl.ac.uk}

\author{Aurora Rossi}
\authornotemark[1]
\correspondingauthor
\orcid{0009-0001-3302-0468}
\affiliation{%
  \institution{University of Bonn}
  \city{Bonn}
  \country{Germany}}
\affiliation{%
  \institution{Lamarr Institute for Machine Learning and Artificial Intelligence}
  \city{Bonn}
  \country{Germany}}
\email{arossi@uni-bonn.de}

\author{Sunil Kumar Maurya}
\orcid{0000-0001-8839-8559}
\affiliation{%
  \institution{The University of Tokyo}
  \city{Tokyo}
  \country{Japan}
}
\email{sunil.maurya@weblab.t.u-tokyo.ac.jp}

\author{Frederik Mallmann-Trenn}
\orcid{0000-0003-0363-8547}
\affiliation{%
  \institution{King's College London}
  \city{London}
  \country{United Kingdom}
}
\email{frederik.mallmann-trenn@kcl.ac.uk}

\author{Xin Liu}
\orcid{0000-0002-2336-7409}
\affiliation{%
  \institution{National Institute of Advanced Industrial Science and Technology}
  \city{Tokyo}
  \country{Japan}
}
\email{xin.liu@aist.go.jp}

\author{Frédéric Giroire}
\orcid{0000-0002-3727-051X}
\affiliation{%
  \institution{Université Côte d’Azur, CNRS, Inria, I3S}
  \city{Sophia Antipolis}
  \country{France}
}
\email{frederic.giroire@inria.fr}

\author{Tsuyoshi Murata}
\orcid{0000-0002-3818-7830}
\affiliation{%
  \institution{Institute of Science Tokyo}
  \city{Tokyo}
  \country{Japan}
}
\email{murata@comp.isct.ac.jp}

\author{Emanuele Natale}
\orcid{0000-0002-8755-3892}
\affiliation{%
  \institution{Université Côte d’Azur, CNRS, Inria, I3S}
  \city{Sophia Antipolis}
  \country{France}
}
\email{emanuele.natale@cnrs.fr}

\renewcommand{\shortauthors}{Justin Dachille et al.}

\begin{abstract}
Computing the importance of nodes in networks is a long-standing fundamental problem that has driven extensive study of various centrality measures.
A particularly well-known centrality measure is betweenness centrality, whose exact computation becomes prohibitive on large-scale networks.
Graph Neural Network (GNN) models have thus been proposed to predict the ranking of nodes by betweenness centrality. However, existing GNN-based methods either have graph-size-dependent parameter counts or are limited to undirected graphs.
We propose a lightweight GNN architecture that exploits the empirically observed relationship between betweenness centrality and multi-hop degree mass. This motivates the use of degree masses as size-invariant node features. To improve generalization, we train on synthetic graphs whose degree distributions more closely match those of real-world networks, including directed and undirected scale-free graphs and uniformly directed hyperbolic random graphs.
We evaluate our model on 14 real-world networks spanning eight domains, including social, email, and citation networks, across both directed and undirected regimes.
The experiments show that our model improves the Kendall $\tau_b$ correlation by up to 24.6\% on undirected and 10.9\% on directed graphs, while using 56$\times$ fewer parameters than the lightest competing GNN baseline and achieving competitive inference time, with up to a
24.5$\times$ speedup on selected directed graphs.

\end{abstract}

\ccsdesc[500]{Computing methodologies~Machine learning}
\ccsdesc[500]{Computing methodologies~Learning to rank}
\ccsdesc[500]{Mathematics of computing~Graph algorithms}
\ccsdesc[500]{Computing methodologies~Neural networks}

\keywords{graph neural network; betweenness centrality; large-scale graph analysis; hyperbolic graph model; degree mass}

\maketitle

\section{Introduction}
Identifying important nodes in networks is a crucial task in many applications. For instance, in social networks, influential individuals can significantly impact information dissemination, while in email communication networks, key individuals often act as hubs for information exchange and organizational coordination \cite{lee_2021,Behera_2019}.
Centrality measures formalize this concept.
In particular, betweenness centrality is effective in identifying nodes that act as bridges or bottlenecks within a network, as it quantifies the importance of a node based on its position in the network by measuring the fraction of shortest paths between pairs of nodes that pass through a given node \cite{Freeman_1977}.

Despite its utility, computing betweenness centrality exactly is computationally expensive: Brandes' algorithm runs in $\mathcal{O}(|V||E|)$ time, making it impractical for large-scale real-world applications \cite{Brandes01062001}.
This has motivated the investigation of approximation methods, including sampling-based algorithms, parallel and distributed computing, and, more recently, learning-based methods such as Graph Neural Networks.
In many downstream applications, the exact centrality value is less important than the induced ranking of nodes, making betweenness approximation naturally suited to learning-to-rank formulations.
Existing GNN-based methods, however, still face important limitations. Some methods rely on node-specific embeddings or padded graph representations, causing the number of learnable parameters to depend on the input graph size. Other parameter-efficient architectures are designed primarily for undirected graphs and do not directly model the asymmetric flow of information in directed networks. This leaves a gap for a compact, size-invariant architecture that can approximate betweenness rankings across both directed and undirected graphs.
We propose BRAVA-GNN (Betweenness Ranking Approximation via Degree Mass GNN)\footnote{Brava means  ``good'' or ``excellent'' in Italian, alluding to the model’s performance.}, a lightweight architecture for betweenness-centrality ranking.
It is designed to handle both directed and undirected graphs, while enhancing accuracy and computational efficiency and maintaining a parameter count independent of graph size.
In particular:
\begin{itemize}
    \item We propose BRAVA-GNN, a compact message-passing architecture for betweenness-centrality ranking that uses multi-hop degree-mass features instead of node-specific embeddings. This makes the number of learnable parameters independent of the input graph size and reduces the parameter count by $56\times$ compared to the lightest GNN baseline.

    \item We introduce a synthetic training strategy based on directed and undirected scale-free graphs and uniformly directed hyperbolic random graphs (UDHY), with generation parameters chosen from empirical real-world network statistics. Our ablations show that this mixed training distribution improves generalization across graph domains.

    \item We evaluate BRAVA-GNN on 14 real-world networks spanning eight domains and compare it against learning-based and sampling-based baselines.
    BRAVA-GNN is best on all test graphs.
    Against the strongest baseline, its gains reach 24.6\% on undirected and 10.9\% on directed graphs, with comparable inference times on undirected graphs, a median of 2.5$\times$ and a peak of $24.5\times$ speedups on directed graphs.
\end{itemize}

The source code is available at \url{https://github.com/justindachille/BRAVA-GNN}.

\section{Related Work}

The exact computation of betweenness centrality relies on Brandes' algorithm \cite{Brandes01062001}, which becomes prohibitively expensive for large-scale graphs due to its time complexity of $\mathcal{O}(|V||E|)$ for unweighted networks and $\mathcal{O}(|V||E| + |V|^2 \log(|V|))$ for weighted networks. Under the Strong Exponential Time Hypothesis \cite{IMPAGLIAZZO2001512}, no truly subquadratic algorithm exists for computing betweenness centrality, as this would imply a subquadratic solver for \texttt{k-SAT*} \cite{BORASSI201651}.
Consequently, there is considerable interest in distributed computing approaches and approximation algorithms.

\subsection{Unsupervised Approximation Algorithms}
\citet{Geisberger2008} improve pivot-sampling approximations of betweenness centrality by rescaling the dependency contributions obtained from randomly selected source nodes. Their distance-aware linear and bisection scaling schemes reduce the overestimation of nodes close to sampled pivots and improve empirical accuracy over the \citet{brandes2007centrality} estimator.
\citet{Riondato_2014} propose a sampling-based approach that leverages the concept of VC-dimension to determine the number of shortest paths required to achieve a desired level of accuracy.
\citet{Borassi_2019} propose KADABRA, which combines balanced bidirectional BFS with adaptive sampling to approximate betweenness centrality in directed and undirected graphs. For fixed additive-error approximation, they report average speedups of more than $100\times$ over previous approximation algorithms on graphs with approximately 10,000 nodes.
\citet{pellegrina_2023} propose SILVAN, a progressive sampling algorithm that uses Monte Carlo Empirical Rademacher Averages with a novel variance-dependent partitioning scheme to obtain sharper non-uniform bounds on the betweenness centrality estimates.
\citet{Mahmoody_2016} address the centrality maximization problem via a randomized algorithm based on shortest-path sampling that provides theoretical approximation guarantees.
\citet{Yoshida_2014} uses hypergraph sketches to compute adaptive betweenness centrality, updating scores in almost linear time as nodes are selected.
Regarding centrality variants, \citet{zhangEfficientApproximationAlgorithms2023} combine deterministic graph traversals with random walk sampling to approximate spanning centrality, while \citet{Brunelli_2024} introduce polynomial-time algorithms for computing betweenness in temporal graphs under various path optimality and waiting constraints.

\subsection{Parallel and Distributed Algorithms}
\citet{AlGhamdi_2017} compute the exact betweenness centrality by implementing a parallel version of Brandes’ algorithm, executed on up to 96,000 CPU cores across multiple supercomputing systems. 
\citet{Bernaschi_2016} address the exact computation on multi-GPU systems by proposing a two-dimensional graph decomposition strategy. \citet{Vella_2018} subsequently extend this line of research with the Dynamic Merging of Frontiers technique, which optimizes graph traversal by unifying exploration frontiers during BFS visits to avoid redundant computations, thereby improving performance on heterogeneous architectures. In parallel, within the context of large-scale distributed systems, \citet{Solomonik_2017} reformulate betweenness computation using linear algebra primitives, presenting an algorithm based on communication-efficient Sparse Matrix Multiplication. 
\citet{Hua_2016} are the first to propose a nearly optimal distributed algorithm for betweenness centrality under the CONGEST model, a standard message-passing abstraction where communication is restricted by limiting the size of messages sent across each edge to $\mathcal{O}(\log n)$ bits per round. \citet{Hoang_2019} extend this line of work with the first nontrivial CONGEST algorithm for betweenness centrality on directed graphs, computing it together with all-pairs shortest paths in $\mathcal{O}(n)$ rounds.

\subsection{Graph Neural Networks}
Graph neural networks have enabled the development of several learning-based methods for approximating centrality measures. Among the earliest GNN-based methods for betweenness-centrality approximation are those of \citet{Maurya_2019} and \citet{fan2019learning}, which appeared concurrently.
\citet{Maurya_2019} introduced GNN-Bet to approximate betweenness centrality, notably by pruning from the adjacency matrix those nodes that lie on no shortest path and by using graph convolutional layers while excluding the node's own feature vector during aggregation. This approach was later extended to closeness centrality in the journal version of the paper \citep{Maurya_2021}.
\citet{fan2019learning} proposed DrBC, a GNN combined with a gated recurrent unit (GRU). This work was further improved by \citet{mirakyan_2021}, who proposed ABCDE, a deep graph convolutional network with a regularization scheme that progressively drops random edges in each convolutional block. Despite its depth, ABCDE uses fewer parameters than DrBC and improves performance in ranking nodes by betweenness centrality. Notably, ABCDE was not compared with \citet{Maurya_2019}.
 We provide a detailed architectural and experimental comparison with these models in Section~\ref{sec:baselines}. At a high level, BRAVA-GNN differs from them by replacing node-specific or single-hop degree features with multi-hop degree-mass features, maintaining a graph-size-independent parameterization, and supporting both directed and undirected graphs.

Beyond static graphs, recent work has also explored GNN-based centrality prediction in dynamic or temporal settings. \citet{Heeg_2024} use De Bruijn graph neural networks to predict temporal centrality measures, including temporal betweenness.

\section{Preliminaries}
This section introduces the notation and background used throughout the paper: betweenness centrality, the node-ranking objective, multi-hop degree mass and the hyperbolic random graph model used to construct part of our synthetic training distribution.
\subsection{Betweenness Centrality}
Given an unweighted graph $G=(V,E)$, either directed or undirected, the betweenness centrality $C_B(v)$ of a node $v \in V$ is defined as:
\begin{equation}
C_B(v) = \sum_{\substack{s,t \in V \setminus \{v\}\\ s \ne t\\ \sigma_{st}>0}} \frac{\sigma_{st}(v)}{\sigma_{st}},
\end{equation}
where $\sigma_{st}$ is the number of shortest paths from node $s$ to node $t$, and $\sigma_{st}(v)$ is the number of those paths that pass through node $v$.
Pairs with no path from $s$ to $t$ are excluded from the sum and therefore contribute zero.
In directed graphs, shortest paths are computed with respect to edge orientation; in undirected graphs, paths may traverse edges in either direction.

\subsection{Problem Definition}
\label{sec:problem}
Given an unweighted graph $G=(V,E)$, either directed or undirected, our goal is to learn a scoring function $f: V \rightarrow \mathbb{R}$ that assigns each node $v \in V$ a real-valued score $f(v)$. These scores induce a node ranking in descending order, where higher-ranked nodes are predicted to have higher betweenness centrality. The target ordering sorts nodes in decreasing $C_B(v)$, so that nodes with larger betweenness centrality appear earlier.

We formulate the task as a ranking problem rather than a regression problem: the objective is not to predict the absolute values of $C_B(v)$, but to preserve the relative ordering of nodes. Accordingly, model quality is evaluated by the Kendall $\tau_b$ correlation between the predicted ranking and the ground-truth betweenness ranking, which accounts for ties in centrality values. This formulation reflects many downstream uses of betweenness centrality, where identifying the most central nodes is more important than estimating their exact centrality scores.

\subsection{Multi-hop Degree Mass}
\label{sec:degree-mass}
The $m$-th order degree mass of a node in a graph $G=(V,E)$ captures the cumulative connectivity around that node up to distance $m$. Originally introduced by \citet{Li_2014}, it is defined as
\begin{equation}
\mathbf{d}^{(m)} = \left( \sum_{k=0}^m A^k \right) \mathbf{d},
\end{equation}
where $A$ is the adjacency matrix and $\mathbf{d}$ is the degree vector.
For directed graphs, degree masses are computed with respect to the matrix supplied to the feature extractor, using that matrix's row-degree vector.
BRAVA-GNN applies this definition to both $A$ and $A^T$ to encode the two edge orientations.

For a node $i$, the $i$-th entry of $\mathbf{d}^{(m)}$ can be written as
\begin{equation}
d_i^{(m)} = \sum_{k=0}^m \sum_{j \in V} (A^k)_{ij} d_j .
\end{equation}
Thus, in an unweighted graph, $(A^k)_{ij}$ counts the number of walks of length $k$ from $i$ to $j$, and $d_i^{(m)}$ aggregates the degrees of nodes reachable from $i$ through walks of length at most $m$, weighted by the number of such walks. The resulting values are nonnegative integers for unweighted graphs.
This quantity is closely related to multi-hop walk-based notions of node influence: nodes with large degree mass are connected, through short walks, to many high-degree nodes. Unlike random-walk-based embeddings or influence estimators, however, degree mass is a deterministic structural feature that can be computed by a small number of sparse matrix-vector multiplications. 
\citet{Li_2014} empirically demonstrate a strong correlation between higher-order degree masses and betweenness centrality in real-world networks, motivating their use as input features for betweenness-ranking prediction.

\subsection{Hyperbolic Random Graph Model}
\label{sec:hyperbolic}
We use hyperbolic random graphs as part of our synthetic training distribution, because they reproduce structural properties commonly observed in real-world networks, including heterogeneous degree distributions, high clustering, and hierarchical organization. The model, introduced by \citet{Krioukov_2010}, samples nodes in a hyperbolic disk, where radial coordinates are associated with node popularity or expected degree, and angular coordinates encode similarity. Edges are then formed preferentially between nodes that are close in the hyperbolic geometry.

In this model, the degree distribution follows a power law $P(k) \sim k^{-\gamma}$, with an exponent controlled by the geometry of the node distribution. In the zero-temperature formulation of \citet{Krioukov_2010}, if the curvature is $K=-\zeta^2$ and $\alpha$ controls the radial density, then
\begin{equation}
    \gamma = \frac{2\alpha}{\zeta} + 1.
\end{equation}
Here, $\zeta>0$ is the curvature scale and $\alpha>0$ controls how strongly nodes concentrate near the boundary of the disk.

In our experiments, we use NetworKit's \texttt{HyperbolicGenerator} \cite{Staudt_2015}, which implements the subquadratic generator of \citet{VonLooz_2016}.
This implementation exposes the model through four parameters: the number of nodes $N$, the average degree $\bar{k}$, the power-law exponent $\gamma$, and the temperature $T$, which controls clustering by smoothing the edge-formation rule.
We therefore treat $\bar{k}$, $\gamma$, and $T$ as the relevant generation parameters for constructing synthetic training graphs.
As described in Section~\ref{sec:trainingset}, $\bar{k}$ and $\gamma$ are sampled from empirical real-world network statistics rather than chosen arbitrarily.

\section{Proposed Method}
BRAVA-GNN first simplifies the input graph with a shortest-path preprocessing heuristic, and then predicts betweenness rankings with a lightweight message-passing architecture based on multi-hop degree-mass features.
\subsection{Preprocessing}
We adopt the preprocessing heuristic introduced in \citet{Maurya_2019}.
For undirected graphs, this rule removes nodes that are guaranteed not to lie on any shortest path.
In particular, nodes satisfying either of the following two conditions are removed:
\begin{enumerate}
    \item \textbf{Isolated or Leaf Nodes:} Any node with fewer than two neighbors has zero centrality, as betweenness is strictly defined for intermediate points.
    \item \textbf{Clique Neighborhoods:} If a node’s neighbors form a clique, any shortest path between them will favor the direct edge (length 1) over a path through the node (length 2).
\end{enumerate}
For directed inputs, these two conditions are evaluated on the underlying undirected graph obtained by ignoring edge orientation.
The retained graph is then the directed subgraph induced by the remaining vertices, so directed shortest paths, degree masses, and message passing still use the directed adjacency.
Thus, directionality is ignored only for the pruning heuristic, and the clique-neighborhood rule is not used as a directed shortest-path certificate.
Observe that, while removed nodes have zero betweenness centrality in the undirected case, removing them can affect the betweenness values of the remaining nodes.
In practice, this impact is negligible compared to the obtained simplification of the graph topology.

\subsection{Architecture}
\label{sec:architecture}
\begin{figure}[t]
    \centering
    \includegraphics[width=0.40\textwidth]{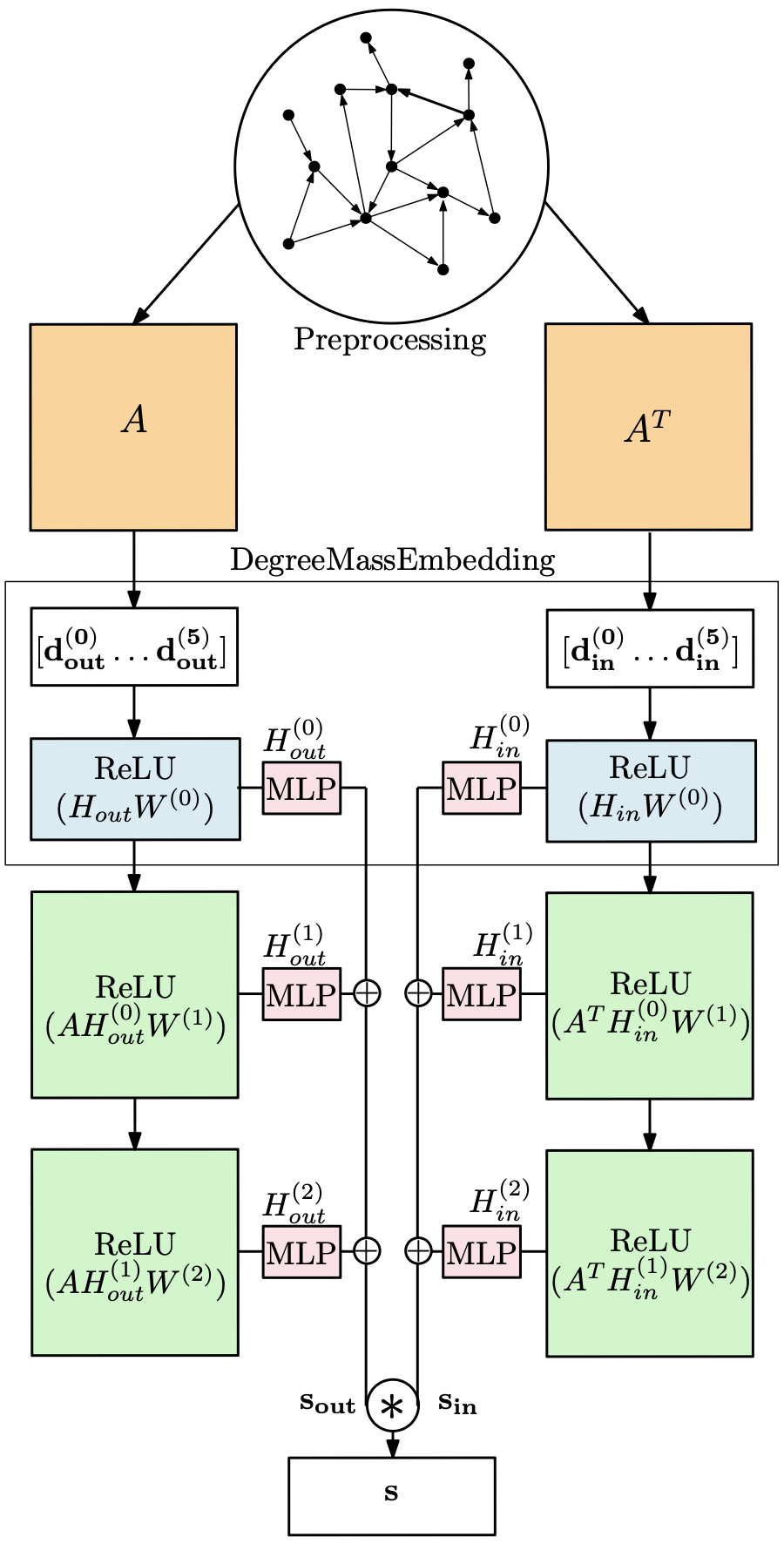}%
    \Description{Architecture diagram of BRAVA-GNN showing degree-mass features, outgoing and incoming message-passing streams, fusion, and scalar node-score prediction.}
    \caption{BRAVA-GNN Architecture Overview. The model utilizes degree masses up to the fifth order as input features, followed by two message-passing layers with ReLU activation and shared MLPs. }
    \label{fig:architecture}
\end{figure}

BRAVA-GNN is a lightweight message-passing architecture designed to produce one scalar score per node. The model uses fixed-dimensional degree-mass features and two directional message-passing streams, allowing the same architecture to operate on graphs of arbitrary size and on both directed and undirected inputs. The complete procedure is summarized in Algorithm~\ref{alg:gnn_ranking} and illustrated in Figure~\ref{fig:architecture}.

\subsubsection{Input Features}
As initial node features, we use the higher-order degree masses introduced in Section~\ref{sec:degree-mass}. For each node, we concatenate degree masses up to order five,
\[
[\mathbf{d}^{(0)}, \mathbf{d}^{(1)}, \ldots, \mathbf{d}^{(5)}],
\]
and map the resulting six-dimensional vector to a hidden representation through a dense layer with ReLU activation. Since the feature dimension is fixed, the number of learnable parameters is independent of the number of nodes in the input graph.

\subsubsection{Dual Message Passing}
In directed graphs, betweenness centrality is determined by directed shortest paths.
A node is central when it can connect sources that reach it to destinations reachable from it.
To capture these two roles, BRAVA-GNN applies two parallel message-passing streams: an outgoing stream over $A$ and an incoming stream over $A^T$.
The two streams share weights, which keeps the architecture compact while applying the same aggregation rule to both directions.

Each stream consists of two message-passing layers. At each layer, a node aggregates hidden representations from its neighbors without adding its own hidden state, following the aggregation scheme used by GNN-Bet (Lines~\ref{alg:gnn_ranking:mp_out}--\ref{alg:gnn_ranking:mp_in}
of Algorithm~\ref{alg:gnn_ranking}). A shared MLP maps the hidden representation at each layer to an intermediate scalar score (Lines~\ref{alg:gnn_ranking:acc_out}--\ref{alg:gnn_ranking:acc_in}
of Algorithm~\ref{alg:gnn_ranking}).

\subsubsection{Score Fusion}
For each direction, intermediate scores from all layers are summed to obtain $s_{out}(v)$ and $s_{in}(v)$. The final score is computed as
\[
s(v) = s_{in}(v) * s_{out}(v).
\]
This multiplicative fusion favors nodes that receive strong evidence from both incoming and outgoing neighborhoods, matching the intuition that high betweenness in directed graphs requires connecting many possible sources to many possible destinations. The fusion strategy is evaluated in the ablation study in Appendix~\ref{sec:ablation}.

\begin{algorithm}[tb]
   \caption{DegreeMassEmbedding.}
   \label{alg:feature_embedding}
\begin{algorithmic}[1]
   \STATE {\bfseries Input:} Adjacency matrix $A$
   \STATE {\bfseries Output:} Initial embedding ${H}$.

   \STATE Compute degree masses $\mathbf{d}^{(0)}, \mathbf{d}^{(1)}, \mathbf{d}^{(2)}, \mathbf{d}^{(3)},\mathbf{d}^{(4)}, \mathbf{d}^{(5)}$ from $A$ (Section~\ref{sec:degree-mass})
   \STATE $ {H}\leftarrow [\mathbf{d}^{(0)}, \mathbf{d}^{(1)}, \mathbf{d}^{(2)}, \mathbf{d}^{(3)}, \mathbf{d}^{(4)}, \mathbf{d}^{(5)}]$ \label{alg:feature_embedding:embedding}
\STATE ${H} \leftarrow \text{ReLU}(H\cdot W^{(0)}_{6\rightarrow 12})$
    \STATE \textbf{return} ${H}$
    
\end{algorithmic}
\end{algorithm}

\begin{algorithm}[tb]
   \caption{BRAVA-GNN.}
   \label{alg:gnn_ranking}
\begin{algorithmic}[1]
   \STATE {\bfseries Input:} Adjacency Matrix $A$, Transpose Adjacency Matrix $A^T$, number of message-passing layers $L$, Weight Matrix $W^{(l)}$ at layer $l$.
   \STATE {\bfseries Output:} Betweenness score $\mathbf{s}$.
   \STATE {\bfseries Notation:} $\operatorname{Norm}_2$ denotes row-wise $L_2$ normalization.
   
    \STATE $H_{out} \leftarrow \text{DegreeMassEmbedding}(A)$
    \STATE $H_{in} \leftarrow \text{DegreeMassEmbedding}(A^T)$
    \STATE $H_{out}^{(0)} \leftarrow\operatorname{Norm}_2\!\left(H_{out} \right) $
    \STATE $H_{in}^{(0)}\leftarrow\operatorname{Norm}_2\!\left(H_{in} \right) $
   
   \STATE $\mathbf{s_{out}} \leftarrow \text{MLP}(H_{out}^{(0)})$
   \STATE $\mathbf{s_{in}} \leftarrow \text{MLP}(H_{in}^{(0)})$
   
   \FOR{each layer $l \in \{0, \dots, L-1\}$}
       \STATE $H_{out}^{(l+1)} \leftarrow \operatorname{Norm}_2\!\left(\text{ReLU}( A \cdot H_{out}^{(l)} \cdot W^{(l+1)} )\right)$ \label{alg:gnn_ranking:mp_out}
       \STATE $H_{in}^{(l+1)} \leftarrow \operatorname{Norm}_2\!\left(\text{ReLU}( A^T \cdot H_{in}^{(l)} \cdot W^{(l+1)} )\right)$ \label{alg:gnn_ranking:mp_in}

       \STATE $\mathbf{s_{out}} \leftarrow \mathbf{s_{out}} + \text{MLP}(H_{out}^{(l+1)})$ \label{alg:gnn_ranking:acc_out}
       \STATE $\mathbf{s_{in}} \leftarrow \mathbf{s_{in}} +\text{MLP}(H_{in}^{(l+1)})$ \label{alg:gnn_ranking:acc_in}
   \ENDFOR
   
   \STATE \textbf{return} $\mathbf{s_{in}} * \mathbf{s_{out}}$
\end{algorithmic}
\end{algorithm}

\section{Experimental Setup}
This section describes the experimental protocol for evaluating BRAVA-GNN, including the training and test datasets, baselines, evaluation metric, loss function, implementation details, and hardware/software environment.

\subsection{Datasets}

\subsubsection{Training set}
\label{sec:trainingset}
To promote generalization across diverse network topologies, we train BRAVA-GNN on a hybrid synthetic dataset composed of directed and undirected scale-free graphs plus uniformly directed hyperbolic random graphs (UDHY).

We include directed scale-free graphs following the protocol of \citet{Maurya_2021}, which has been widely used in previous GNN-based work on betweenness approximation. However, scale-free graphs alone capture only part of the structural variability observed in real networks. We therefore complement them with uniformly directed versions of hyperbolic random graphs, described in Section~\ref{sec:hyperbolic}, which can reproduce several properties commonly found in real-world networks, including heterogeneous degree distributions, high clustering coefficients, and small-world structure \cite{Gugelmann_2012,Friedrich_2018,MULLER_2019}.

Rather than selecting generation parameters arbitrarily, we adopt a data-driven parameterization strategy. We analyze a reference collection of 10 real-world networks \cite{Network_Data_Repo,snapnets} to estimate empirical values of the average degree $\bar{k}$ and the power-law exponent $\gamma$; these statistics are reported in Table~\ref{tab:hyperbolic_params}. These empirical distributions are then used to parameterize the synthetic hyperbolic graphs.

Our final training set consists of 30 synthetic graphs, each with $N=100{,}000$ nodes:
\begin{itemize}
    \item \textbf{10 Directed Scale-Free Graphs:} generated using the configuration detailed in \citet{Maurya_2021}.
    \item \textbf{10 Undirected Scale-Free Graphs:} generated using the same configuration and then symmetrized by replacing each directed edge $u \to v$ with the undirected pair $\{u,v\}$.
    \item \textbf{10 Directed Hyperbolic Random Graphs:} generated by sampling the temperature $T$ uniformly from $(0,0.5)$, while $\bar{k}$ and $\gamma$ are drawn from the empirical distributions in Table~\ref{tab:hyperbolic_params}. Each undirected edge $\{u,v\}$ is then oriented uniformly at random as either $u \to v$ or $v \to u$.
\end{itemize}
For each synthetic graph, exact betweenness centrality scores are computed and used to derive the ground-truth ranking labels.

\subsubsection{Test set}
For evaluation, we use 14 real-world networks from the Network Data Repository \cite{Network_Data_Repo} and SNAP \cite{snapnets}. The benchmark spans eight domains: social, web, wiki, email, peer-to-peer, citation, authorship, and co-purchase networks. It includes both directed and undirected graphs and covers a broad range of graph sizes, from peer-to-peer networks with tens of thousands of nodes to large-scale graphs with millions of nodes and edges.

To ensure comparability with prior work, our test set includes the graphs used by \citet{Maurya_2021}, as well as the subset used by \citet{mirakyan_2021} and \citet{fan2019learning}: \texttt{com-youtube}, \texttt{amazon}, \texttt{dblp}, \texttt{cit-Patents}, and \texttt{com-lj}. Detailed node and edge counts for all test graphs are reported in Table~\ref{tab:datasets} in the Appendix.

\subsection{Baselines}
\label{sec:baselines}
We compare BRAVA-GNN with two families of baselines.
The learning baselines are \textbf{GNN-Bet}, \textbf{DrBC}, and \textbf{ABCDE} \cite{Maurya_2019,fan2019learning,mirakyan_2021}.
The sampling baselines are \textbf{KADABRA}, \textbf{SILVAN}, and \textbf{Bavarian} \cite{Borassi_2019,pellegrina_2023,Cousins_2023_Bavarian}.
ABCDE has previously been shown to outperform earlier sampling-based methods such as ABRA \cite{10.1145/3208351} and RK \cite{Riondato_2014}, as well as the learning-based Node2Vec baseline \cite{10.1145/2939672.2939754}; we therefore omit these earlier methods from direct comparison.

\subsubsection{GNN-Bet}
GNN-Bet \cite{Maurya_2019} is the closest baseline to our work. We adopt its preprocessing heuristic and build on its idea of processing both edge directions. However, GNN-Bet uses learnable node-specific embeddings, making its number of parameters depend on the graph size and requiring padding to the largest graph in the training set. BRAVA-GNN instead uses fixed-dimensional degree-mass features, reducing the parameter count from $\approx$\,58M to $\approx$\,1.3k and allowing the same model to be applied to graphs of different sizes.

\subsubsection{ABCDE}
ABCDE \cite{mirakyan_2021} represents a significant step towards parameter-efficient centrality approximation. It uses a deep graph convolutional architecture regularized by \textit{Progressive-DropEdge}, which progressively drops random edges in each convolutional block to reduce overfitting and over-smoothing. ABCDE uses node degree as its input feature and relies on multiple convolutional layers to aggregate wider neighborhood information. Compared with ABCDE, BRAVA-GNN uses higher-order degree-mass features directly and has a smaller parameter count ($\approx$\,1.3k vs.\ $\approx$\,72k). ABCDE is designed for undirected graphs, whereas BRAVA-GNN is evaluated on both directed and undirected networks.

\subsubsection{DrBC}
DrBC \cite{fan2019learning} is a GNN-based model for betweenness-centrality ranking that uses degree-based initial node features. It aggregates information from neighboring nodes and updates node representations through a gated recurrent unit (GRU). To combine information from different propagation depths, DrBC applies max-pooling over the representations produced at each layer. Like ABCDE, DrBC is designed for undirected graphs.

\subsubsection{KADABRA}
KADABRA \cite{Borassi_2019} is a randomized sampling algorithm for approximating betweenness centrality in directed and undirected graphs. It repeatedly samples shortest paths between pairs of nodes and updates the estimates of the nodes lying on the sampled paths. Its adaptive stopping criterion uses probabilistic confidence intervals to stop once the target approximation accuracy is reached.
We use an error tolerance of $\epsilon=0.01$ and a failure probability of $\delta=0.1$, matching the configuration used for the other sampling-based baselines when applicable.
\subsubsection{SILVAN}
SILVAN \cite{pellegrina_2023} is a randomized shortest-path sampling algorithm for betweenness centrality approximation with probabilistic guarantees. Unlike KADABRA, it uses a progressive sampling scheme based on Monte Carlo Empirical Rademacher Averages and non-uniform bounds, refining the confidence estimates in a node-dependent way. This can reduce the number of samples needed to obtain high-quality approximations.
We set the approximation accuracy to $\epsilon=0.01$ and the failure probability to $\delta=0.05$, following the default confidence setting used in the original SILVAN implementation.

\subsubsection{Bavarian}
Bavarian \cite{Cousins_2023_Bavarian} is a sampling-based algorithmic framework for betweenness centrality approximation. It combines shortest-path sampling with Monte Carlo Empirical Rademacher Averages and variance-aware probabilistic bounds to obtain tight, data-dependent approximation guarantees.
We use the RK estimator variant (Riondato--Kornaropoulos \cite{Riondato_2014}), with $\epsilon=0.01$, $\delta=0.1$, 100 Monte Carlo trials, and a progressive scaling factor of 2.0.

\begin{table}[t]
  \centering
  \caption{Reference networks for estimating $\gamma$, $\bar{k}$, and $\rho$ (degree-betweenness correlation); dash = not computed.}
  \label{tab:hyperbolic_params}
  \begin{tabular}{@{}lrrrrr@{}}
    \toprule
    \textbf{Graph} & \textbf{\#Nodes} & \textbf{\#Edges} & $\bar{k}$ & $\gamma$ & $\rho$ \\
    \midrule
    p2p-Gnutella30 & 36,682 & 88,328 & 4.8159 & 7.0192 & 0.9287 \\
    email-Enron & 36,692 & 183,831 & 10.0202 & 2.3982 & 0.8157 \\
    musae-github & 37,700 & 289,003 & 15.3317 & 2.5023 & 0.8884 \\
    soc-Slashdot0811 & 77,360 & 828,161 & 21.4106 & 2.1604 & 0.8442 \\
    gemsec-Facebook & 50,515 & 819,090 & 32.4296 & 2.4494 & 0.7246 \\
    twitch-gamers & 168,114 & 6,797,557 & 80.8684 & 2.1334 & 0.7553 \\
    com-DBLP & 317,080 & 1,049,866 & 6.6221 & 3.5380 & 0.7200 \\
    web-NotreDame & 325,729 & 1,469,679 & 9.0239 & 2.0757 & 0.2905 \\
    web-BerkStan & 685,230 & 7,600,595 & 22.1841 & 2.6245 & 0.4503 \\
    com-Orkut & 3,072,441 & 117,185,083 & 76.2814 & 2.2849 & -- \\
    \bottomrule
  \end{tabular}
\end{table}

\begin{table*}
  \caption{Ranking accuracy (Kendall $\tau_b$, $\uparrow$; mean $\pm$ std.\ over 3 seeds). Best/second-best are bolded/underlined. \% Imp.\ is BRAVA-GNN's gain over the strongest baseline. Dash = unavailable: ABCDE/DrBC are undirected-only; Bavarian ran out of memory.}
  \label{tab:main_accuracy}
  \centering
  \small
  \begin{tabular*}{\textwidth}{@{\extracolsep{\fill}}lllllllll@{}}
    \toprule
    \textbf{Graph} & \textbf{GNN-Bet} & \textbf{ABCDE} & \textbf{DrBC} & \textbf{KADABRA} & \textbf{SILVAN} & \textbf{Bavarian} & \textbf{BRAVA-GNN} & \textbf{\% Imp.} \\
    \midrule
    \multicolumn{9}{@{}l@{}}{\textbf{\textit{Undirected}}} \\
    \hspace{0.5em}p2p-Gnutella31 & $\underline{89.3 \pm 1.1}$ & $72.6 \pm 0.8$ & $78.2 \pm 0.0$ & $72.0 \pm 0.0$ & $66.9 \pm 1.1$ & $68.8 \pm 0.1$ & $\mathbf{89.7 \pm 0.7}$ & +0.4\% \\
    \hspace{0.5em}com-youtube & $\underline{76.3 \pm 0.2}$ & $56.2 \pm 0.3$ & $49.5 \pm 0.0$ & $32.3 \pm 0.1$ & $44.7 \pm 0.0$ & $49.0 \pm 5.5$ & $\mathbf{92.2 \pm 0.1}$ & +20.9\% \\
    \hspace{0.5em}amazon & $\underline{68.9 \pm 0.2}$ & $65.9 \pm 1.2$ & $66.2 \pm 0.0$ & $19.0 \pm 0.1$ & $24.7 \pm 0.3$ & $20.4 \pm 0.1$ & $\mathbf{85.9 \pm 1.7}$ & +24.6\% \\
    \hspace{0.5em}cit-Patents & $57.5 \pm 0.2$ & $66.7 \pm 1.0$ & $\underline{68.2 \pm 0.0}$ & $18.3 \pm 0.0$ & $17.9 \pm 0.2$ & $15.4 \pm 0.0$ & $\mathbf{74.0 \pm 0.2}$ & +8.6\% \\
    \hspace{0.5em}com-lj & $\underline{68.8 \pm 0.4}$ & $65.3 \pm 1.2$ & $62.2 \pm 0.0$ & $16.7 \pm 0.0$ & $28.9 \pm 0.0$ & -- & $\mathbf{80.5 \pm 0.2}$ & +17.0\% \\
    \hspace{0.5em}dblp & $76.2 \pm 0.7$ & $\underline{78.6 \pm 0.6}$ & $69.1 \pm 0.0$ & $18.1 \pm 0.0$ & $17.2 \pm 1.1$ & -- & $\mathbf{85.0 \pm 1.3}$ & +8.2\% \\
    \textbf{Undirected AVG} & $\underline{72.8}$ & $67.6$ & $65.6$ & $29.4$ & $33.4$ & -- & $\mathbf{84.6}$ & +16.1\% \\
    \cmidrule(lr){1-9}
    \multicolumn{9}{@{}l@{}}{\textbf{\textit{Directed}}} \\
    \hspace{0.5em}soc-Epinions1 & $\underline{87.9 \pm 0.2}$ & -- & -- & $53.7 \pm 0.0$ & $49.0 \pm 0.5$ & $56.6 \pm 0.2$ & $\mathbf{92.6 \pm 0.1}$ & +5.3\% \\
    \hspace{0.5em}soc-Slashdot0902 & $\underline{81.4 \pm 0.6}$ & -- & -- & $51.9 \pm 0.0$ & $55.8 \pm 0.1$ & $61.6 \pm 0.0$ & $\mathbf{90.3 \pm 0.4}$ & +10.9\% \\
    \hspace{0.5em}email-EuAll & $\underline{98.9 \pm 0.1}$ & -- & -- & $31.7 \pm 0.2$ & $32.2 \pm 1.1$ & $30.0 \pm 0.2$ & $\mathbf{99.2 \pm 0.0}$ & +0.3\% \\
    \hspace{0.5em}web-Google & $\underline{77.6 \pm 0.2}$ & -- & -- & $34.0 \pm 0.0$ & $30.6 \pm 0.1$ & $39.2 \pm 0.2$ & $\mathbf{79.8 \pm 0.4}$ & +2.7\% \\
    \hspace{0.5em}soc-Pokec & $\underline{76.0 \pm 0.3}$ & -- & -- & $22.2 \pm 0.3$ & $35.8 \pm 0.2$ & -- & $\mathbf{83.3 \pm 0.6}$ & +9.6\% \\
    \hspace{0.5em}wiki-topcats & $\underline{58.4 \pm 1.5}$ & -- & -- & $21.5 \pm 0.0$ & $28.4 \pm 0.4$ & $24.4 \pm 0.1$ & $\mathbf{63.1 \pm 0.9}$ & +8.1\% \\
    \hspace{0.5em}wiki-Talk & $\underline{98.8 \pm 0.1}$ & -- & -- & $12.1 \pm 0.0$ & $5.8 \pm 0.2$ & $8.4 \pm 0.3$ & $\mathbf{99.3 \pm 0.0}$ & +0.5\% \\
    \hspace{0.5em}soc-LiveJournal1 & $\underline{75.9 \pm 0.3}$ & -- & -- & $14.7 \pm 0.0$ & $20.5 \pm 0.8$ & -- & $\mathbf{82.8 \pm 0.2}$ & +9.0\% \\
    \textbf{Directed AVG} & $\underline{81.9}$ & -- & -- & $30.2$ & $32.3$ & -- & $\mathbf{86.3}$ & +5.4\% \\
    \midrule
    \textbf{AVG} & $\underline{78.0}$ & -- & -- & $29.9$ & $32.8$ & -- & $\mathbf{85.5}$ & +9.7\% \\
    \bottomrule
  \end{tabular*}
\end{table*}

\begin{table*}[t]
  \caption{Inference time (seconds, $\downarrow$; mean $\pm$ std.\ over 3 seeds). Brandes is exact (single run). Best/second-best are bolded/underlined. AVG speedup is the median per-graph best-baseline-to-BRAVA ratio. Dash = unavailable: ABCDE/DrBC are undirected-only; Bavarian ran out of memory; Brandes timed out after 7d.}
  \label{tab:main_time}
  \centering
  \small
  \begin{tabular*}{\textwidth}{@{\extracolsep{\fill}}llllllllll@{}}
    \toprule
    \textbf{Graph} & \textbf{GNN-Bet} & \textbf{ABCDE} & \textbf{DrBC} & \textbf{KADABRA} & \textbf{SILVAN} & \textbf{Bavarian} & \textbf{Brandes} & \textbf{BRAVA-GNN} & \textbf{Speedup $\uparrow$} \\
    \midrule
    \multicolumn{10}{@{}l@{}}{\textbf{\textit{Undirected}}} \\
    \hspace{0.5em}p2p-Gnutella31 & $0.6 \pm 0.0$ & $0.4 \pm 0.0$ & $\underline{0.2 \pm 0.0}$ & $0.6 \pm 0.1$ & $0.6 \pm 0.1$ & $13.7 \pm 0.9$ & $48.3$ & $\mathbf{<0.05}$ & 13.8$\times$ \\
    \hspace{0.5em}com-youtube & $\underline{0.6 \pm 0.0}$ & $10.7 \pm 0.3$ & $4.5 \pm 0.3$ & $15.8 \pm 0.6$ & $17.1 \pm 2.5$ & $1312.6 \pm 442.4$ & $3416.7$ & $\mathbf{0.3 \pm 0.0}$ & 2.1$\times$ \\
    \hspace{0.5em}amazon & $\mathbf{0.7 \pm 0.0}$ & $24.0 \pm 1.5$ & $11.2 \pm 0.4$ & $28.4 \pm 1.3$ & $35.2 \pm 2.4$ & $990.0 \pm 10.6$ & $14061.1$ & $\underline{0.7 \pm 0.1}$ & 1.0$\times$ \\
    \hspace{0.5em}cit-Patents & $\mathbf{0.9 \pm 0.0}$ & $46.4 \pm 3.1$ & $16.9 \pm 1.1$ & $53.9 \pm 0.2$ & $56.5 \pm 0.9$ & $1216.4 \pm 4.9$ & $23003.0$ & $\underline{2.1 \pm 0.5}$ & 0.4$\times$ \\
    \hspace{0.5em}com-lj & $\mathbf{1.4 \pm 0.0}$ & $88.4 \pm 2.7$ & $26.4 \pm 0.5$ & $88.9 \pm 1.5$ & $91.3 \pm 0.5$ & -- & $143715.3$ & $\underline{3.8 \pm 0.7}$ & 0.4$\times$ \\
    \hspace{0.5em}dblp & $\mathbf{0.7 \pm 0.0}$ & $28.0 \pm 0.1$ & $11.9 \pm 0.2$ & $59.9 \pm 1.4$ & $58.6 \pm 5.5$ & -- & $15102.6$ & $\underline{1.0 \pm 0.2}$ & 0.6$\times$ \\
    \textbf{Undirected AVG} & $\mathbf{0.8}$ & $33.0$ & $11.9$ & $41.2$ & $43.2$ & -- & $33224.5$ & $\underline{1.3}$ & 0.8$\times$ \\
    \cmidrule(lr){1-10}
    \multicolumn{10}{@{}l@{}}{\textbf{\textit{Directed}}} \\
    \hspace{0.5em}soc-Epinions1 & $\underline{0.6 \pm 0.0}$ & -- & -- & $1.3 \pm 0.2$ & $1.0 \pm 0.0$ & $59.1 \pm 1.7$ & $33.6$ & $\mathbf{<0.05}$ & 24.5$\times$ \\
    \hspace{0.5em}soc-Slashdot0902 & $\underline{0.7 \pm 0.0}$ & -- & -- & $1.7 \pm 0.0$ & $1.8 \pm 0.3$ & $57.5 \pm 23.8$ & $97.7$ & $\mathbf{<0.05}$ & 15.9$\times$ \\
    \hspace{0.5em}email-EuAll & $\underline{0.6 \pm 0.0}$ & -- & -- & $2.4 \pm 0.6$ & $2.1 \pm 0.5$ & $198.7 \pm 1.9$ & $96.4$ & $\mathbf{<0.05}$ & 23.4$\times$ \\
    \hspace{0.5em}web-Google & $\underline{0.7 \pm 0.0}$ & -- & -- & $17.3 \pm 1.3$ & $16.9 \pm 0.4$ & $6322.9 \pm 1032.3$ & $4191.0$ & $\mathbf{0.2 \pm 0.0}$ & 2.9$\times$ \\
    \hspace{0.5em}soc-Pokec & $\mathbf{1.4 \pm 0.0}$ & -- & -- & $82.5 \pm 1.3$ & $83.0 \pm 0.1$ & -- & $80363.8$ & $\underline{1.7 \pm 0.4}$ & 0.8$\times$ \\
    \hspace{0.5em}wiki-topcats & $\mathbf{1.3 \pm 0.0}$ & -- & -- & $61.6 \pm 2.1$ & $61.4 \pm 1.4$ & $2083.2 \pm 702.9$ & $69991.3$ & $\underline{1.5 \pm 0.2}$ & 0.9$\times$ \\
    \hspace{0.5em}wiki-Talk & $\underline{0.6 \pm 0.0}$ & -- & -- & $30.1 \pm 1.8$ & $30.6 \pm 0.7$ & $1207.3 \pm 209.8$ & $9909.6$ & $\mathbf{0.3 \pm 0.1}$ & 2.0$\times$ \\
    \hspace{0.5em}soc-LiveJournal1 & $\mathbf{2.3 \pm 0.0}$ & -- & -- & $161.3 \pm 4.4$ & $163.6 \pm 2.8$ & -- & -- & $\underline{3.8 \pm 0.7}$ & 0.6$\times$ \\
    \textbf{Directed AVG} & $\underline{1.0}$ & -- & -- & $44.8$ & $45.0$ & -- & -- & $\mathbf{1.0}$ & 2.5$\times$ \\
    \midrule
    \textbf{AVG} & $\mathbf{0.9}$ & -- & -- & $43.3$ & $44.3$ & -- & -- & $\underline{1.1}$ & 1.5$\times$ \\
    \bottomrule
  \end{tabular*}
\end{table*}

\subsection{Loss Function}
\label{sec:loss_function}
As motivated in Section~\ref{sec:problem}, we formulate centrality prediction as a pairwise ranking problem rather than regression. We employ a \textit{Margin Ranking Loss} to enforce the correct ordering of nodes based on their centrality scores. Given two nodes $u$ and $v$, the loss is defined as:
\begin{equation}
\mathcal{L}(s_u, s_v, y) = \max(0, 1 -y \cdot (s_u - s_v) ),
\end{equation}
where $s_u$ and $s_v$ are the predicted scores generated by the model for nodes $u$ and $v$, respectively.
The label $y \in \{1, -1\}$ serves as the ranking indicator: we set $y=1$ if $u$ has strictly larger ground-truth betweenness than $v$, and $y=-1$ in the opposite case.
Pairs with equal ground-truth betweenness are omitted from the loss because they do not define a strict preference.
This formulation aligns with the pairwise hinge loss used in Learning to Rank frameworks \cite{Chen_2009}.

\subsection{Evaluation Metric}

To evaluate the performance of our model in the node ranking task, we use Kendall's $\tau_b$ coefficient.
This metric measures the ordinal association between the predicted and the ground-truth rankings.
The $\tau_b$ variant accounts for ties in the rankings, which is particularly relevant for betweenness centrality, where large subsets of nodes (e.g., peripheral nodes or symmetric structures) often share identical scores.
Formally, the Kendall $\tau_b$ coefficient is defined as:
\begin{equation}
\tau_b = \frac{C - D}{\sqrt{(C + D + T)(C + D + U)}},
\end{equation}
where $C$ denotes the number of \textit{concordant pairs} (pairs of nodes $(i, j)$ whose ranks agree in both the predicted and true rankings), $D$ denotes the number of \textit{discordant pairs} (pairs whose ranks are reversed), $T$ is the number of ties only in the predicted ranking, and $U$ is the number of ties only in the true ranking.
The $\tau_b$ coefficient ranges from $-1$ (perfect disagreement) to $1$ (perfect agreement), where a higher score indicates that the model has learned to preserve the topological hierarchy of the network. For readability, all reported $\tau_b$ values are multiplied by $100$.

\subsection{Implementation Details}
\label{sec:model_config}
We train BRAVA-GNN for 10 epochs using the Adam optimizer with learning rate $5 \times 10^{-3}$. 
The final architecture uses $L=2$ message-passing layers and hidden dimension $n_{hid}=12$. The prediction MLP consists of three linear layers with dimensions $12 \rightarrow 24 \rightarrow 24 \rightarrow 1$, ReLU activations, and dropout $p=0.3$. Together with the $6 \rightarrow 12$ degree-mass encoder and the two shared message-passing matrices, this gives 1{,}297 learnable parameters (Table~\ref{tab:model_params}), independent of the input graph size.
The hyperparameters were selected through the ablation studies reported in Appendix~\ref{sec:ablation}, balancing ranking accuracy and parameter efficiency.
Accuracy is flat across the tested depths and widths at $L=2$ (Table~\ref{tab:depth_width}); we adopt $n_{hid}=12$ which reaches $\tau_b = 85.5 \pm 0.5$.
The case $L=0$ corresponds to removing message passing entirely; in this setting, the model reduces to an MLP applied directly to the degree-mass features.
The same ablations show that five-hop degree mass performs best among the tested feature orders (Table~\ref{tab:init}), and that dropout $p=0.3$ gives the best performance among the tested rates $0.0$--$0.5$ (Table~\ref{tab:dropout_fusion}).

\subsection{Hardware and Software}

BRAVA-GNN is implemented in PyTorch \cite{paszke2019pytorch}. We use NetworkX \cite{Hagberg_2008} for graph generation and data processing, and NetworKit \cite{Staudt_2015} to compute exact betweenness centrality values and to create the hyperbolic graphs.

All runs reported in this paper, including training, inference, and wall-clock baseline evaluations, were executed on the same hardware platform: 2$\times$ Intel Xeon Gold 6230R CPUs (52 cores total), 384 GiB of RAM, and 4$\times$ NVIDIA Quadro RTX 8000 GPUs with 45 GiB of usable GPU memory each.
To assess statistical robustness, we repeat training with three random seeds and report mean performance and standard deviation across independent runs. The complete training process takes approximately 18 minutes on a single NVIDIA Quadro RTX 8000 GPU.
Each run used a single GPU, and the machine was reserved exclusively during execution to make wall-clock timings comparable across methods.
Reported inference times measure a single forward pass.
For BRAVA-GNN, they also include the computation of degree-mass features $[\mathbf{d}^{(0)},\dots,\mathbf{d}^{(5)}]$ from the adjacency matrix.

\section{Results}

Tables~\ref{tab:main_accuracy} and~\ref{tab:main_time} compare BRAVA-GNN against the baselines in ranking accuracy (Kendall $\tau_b$) and inference time (seconds), respectively.
Parameter count is reported separately in Table~\ref{tab:model_params}.

BRAVA-GNN attains the best Kendall $\tau_b$ on all 14 test graphs (Table~\ref{tab:main_accuracy}), improving upon the strongest baseline by 9.7\% on average. The gain is consistent across both regimes, with +16.1\% on undirected graphs and +5.4\% on directed graphs, and peak per-graph improvements of +24.6\% and +10.9\% respectively. ABCDE and DrBC cannot process directed adjacencies and are therefore absent from the directed columns, underscoring BRAVA-GNN's broader applicability.

\paragraph{Parameter efficiency.}
Table~\ref{tab:model_params} reports the number of learnable parameters for each model.
With only 1{,}297 parameters, BRAVA-GNN is $56\times$ smaller than ABCDE, $96\times$ smaller than DrBC, and over four orders of magnitude smaller than GNN-Bet, whose embedding-based design ties model size to the largest training graph.
This compactness does not come at the cost of runtime: as shown in Table~\ref{tab:main_time}, BRAVA-GNN's inference latency is competitive with the fastest GNN baseline on every dataset and remains well below the sampling-based methods on the largest graphs.

\begin{table}[ht]
    \centering
    \caption{\textbf{Number of learnable parameters.} GNN-Bet \cite{Maurya_2019} scales with graph size ($|V|$), while DrBC \cite{fan2019learning}, ABCDE \cite{mirakyan_2021}, and our model are size-invariant. Notably, BRAVA-GNN requires $\sim56\times$ fewer parameters than ABCDE.}
    \label{tab:model_params}
    \begin{tabular}{lr}
        \toprule
        \textbf{Model} & \textbf{\# Parameters} \\
        \midrule
        GNN-Bet & 58,172,425 \\
        DrBC & 124,225 \\
        ABCDE & \underline{72,477} \\
        BRAVA-GNN & \textbf{1,297} \\
        \bottomrule
    \end{tabular}
\end{table}

\section{Conclusion}
In this work, we introduced BRAVA-GNN, a lightweight framework for approximating betweenness centrality rankings. The method is based on two key design choices. First, it replaces the node-specific embeddings used by previous GNN-based approaches, whose parameterization may depend on the input graph size, with size-invariant multi-hop degree-mass features. Second, it is also trained on uniformly directed hyperbolic random graphs, whose power-law degree distributions and clustering patterns better reflect the structure of real-world networks than the scale-free synthetic graphs commonly used in prior work.

These choices enable BRAVA-GNN to achieve strong generalization with only about 1{,}297 trainable parameters, substantially fewer than existing GNN-based baselines. Across all 14 test graphs, covering both directed and undirected settings, BRAVA-GNN consistently outperforms larger models in betweenness ranking accuracy.

Future work will investigate whether the same topological learning principle can be extended to other path-based centrality measures, such as closeness centrality. Another promising direction is the adaptation of BRAVA-GNN to temporal graphs, where centrality rankings evolve over time and efficient updates become essential.

\begin{table*}[t]
  \centering
  \scriptsize
  \renewcommand{\arraystretch}{1.15}
  \setlength{\tabcolsep}{2pt}
  \caption{Ablation: Training Distribution (accuracy).}
  \label{tab:train_distribution}
  \resizebox{\textwidth}{!}{%
  \begin{tabular}{@{}l*{15}{c}@{}}
    \toprule
    \textbf{Graph} &
    \textbf{BO} &
    \textbf{SBO} &
    \textbf{UDHY} &
    \textbf{HY} &
    \textbf{-SBO} &
    \textbf{-BO} &
    \textbf{-UDHY} &
    \textbf{-HY$^\dagger$} &
    \textbf{Dir-only} &
    \textbf{Undir-only} &
    \textbf{BO+HY} &
    \textbf{All-4} &
    \textbf{mNN-swap} &
    \textbf{BO+SBO+mNN} &
    \textbf{BA-swap} \\
    \midrule
    \multicolumn{16}{@{}l@{}}{\textbf{\textit{Undirected}}} \\
    \hspace{0.5em}p2p-Gnutella31
    & $90.7 \pm 2.0$ & $91.4 \pm 2.4$ & $89.4 \pm 0.3$ & $88.8 \pm 0.6$
    & $91.1 \pm 0.3$ & $88.9 \pm 3.3$ & $88.4 \pm 4.0$
    & $\underline{93.2 \pm 1.2}$ & $91.8 \pm 0.9$ & $89.7 \pm 1.9$
    & $87.4 \pm 0.1$ & $89.5 \pm 1.4$ & $89.7 \pm 1.0$ & $91.6 \pm 0.7$
    & $\mathbf{94.5 \pm 1.1}$ \\
    \hspace{0.5em}com-youtube
    & $81.9 \pm 1.2$ & $\mathbf{91.8 \pm 0.8}$ & $89.2 \pm 0.3$ & $88.5 \pm 0.8$
    & $90.9 \pm 0.1$ & $89.2 \pm 0.8$ & $91.2 \pm 0.2$
    & $91.0 \pm 0.5$ & $89.3 \pm 0.9$ & $\underline{91.3 \pm 0.8}$
    & $87.8 \pm 0.9$ & $90.8 \pm 0.1$ & $90.2 \pm 0.1$ & $90.2 \pm 0.7$
    & $87.8 \pm 0.4$ \\
    \hspace{0.5em}amazon
    & $75.1 \pm 1.6$ & $\mathbf{86.6 \pm 1.5}$ & $76.7 \pm 0.2$ & $75.7 \pm 1.8$
    & $82.3 \pm 0.8$ & $77.7 \pm 0.8$ & $83.9 \pm 1.1$
    & $82.8 \pm 2.1$ & $78.4 \pm 0.8$ & $\underline{85.1 \pm 1.2}$
    & $77.1 \pm 2.2$ & $82.0 \pm 1.1$ & $81.0 \pm 1.6$ & $85.1 \pm 0.8$
    & $80.8 \pm 1.6$ \\
    \hspace{0.5em}cit-Patents
    & $71.0 \pm 1.1$ & $71.1 \pm 4.0$ & $72.1 \pm 0.7$ & $71.9 \pm 1.5$
    & $\underline{73.8 \pm 0.7}$ & $72.6 \pm 1.5$ & $72.1 \pm 2.0$
    & $72.8 \pm 0.7$ & $\mathbf{74.1 \pm 1.1}$ & $72.3 \pm 0.8$
    & $70.8 \pm 1.0$ & $73.4 \pm 0.5$ & $73.1 \pm 0.2$ & $70.2 \pm 2.0$
    & $72.2 \pm 0.5$ \\
    \hspace{0.5em}com-lj
    & $73.3 \pm 0.6$ & $75.4 \pm 3.4$ & $79.8 \pm 0.6$ & $77.7 \pm 0.6$
    & $\underline{80.2 \pm 0.2}$ & $79.4 \pm 0.3$ & $79.8 \pm 1.0$
    & $79.5 \pm 0.2$ & $79.8 \pm 0.4$ & $\mathbf{80.4 \pm 0.6}$
    & $78.8 \pm 0.7$ & $79.8 \pm 0.7$ & $79.7 \pm 0.0$ & $77.5 \pm 0.4$
    & $75.6 \pm 1.1$ \\
    \hspace{0.5em}dblp
    & $83.9 \pm 3.8$ & $\underline{86.4 \pm 1.1}$ & $83.5 \pm 1.8$ & $83.0 \pm 1.6$
    & $82.8 \pm 1.4$ & $83.4 \pm 1.7$ & $82.9 \pm 1.7$
    & $85.6 \pm 1.7$ & $86.3 \pm 1.7$ & $82.1 \pm 0.7$
    & $79.8 \pm 2.3$ & $84.3 \pm 0.2$ & $80.1 \pm 1.3$ & $82.2 \pm 1.4$
    & $\mathbf{87.1 \pm 1.4}$ \\
    \cmidrule(lr){1-16}
    \multicolumn{16}{@{}l@{}}{\textbf{\textit{Directed}}} \\
    \hspace{0.5em}soc-Epinions1
    & $88.4 \pm 0.2$ & $91.6 \pm 0.4$ & $92.1 \pm 0.3$ & $90.9 \pm 0.4$
    & $\underline{92.4 \pm 0.2}$ & $91.8 \pm 0.3$ & $92.0 \pm 0.3$
    & $\mathbf{92.7 \pm 0.1}$ & $92.1 \pm 0.3$ & $92.1 \pm 0.3$
    & $91.1 \pm 0.3$ & $92.2 \pm 0.2$ & $92.1 \pm 0.1$ & $92.1 \pm 0.2$
    & $91.1 \pm 0.3$ \\
    \hspace{0.5em}soc-Slashdot0902
    & $84.2 \pm 0.5$ & $\underline{89.2 \pm 0.3}$ & $88.4 \pm 0.1$ & $86.9 \pm 0.7$
    & $88.4 \pm 0.5$ & $87.2 \pm 0.4$ & $88.3 \pm 0.3$
    & $\mathbf{89.6 \pm 0.5}$ & $88.7 \pm 0.5$ & $88.4 \pm 0.1$
    & $85.2 \pm 0.9$ & $88.5 \pm 0.3$ & $87.6 \pm 0.9$ & $88.5 \pm 0.6$
    & $87.6 \pm 0.3$ \\
    \hspace{0.5em}email-EuAll
    & $99.0 \pm 0.1$ & $99.1 \pm 0.0$ & $99.1 \pm 0.0$ & $99.0 \pm 0.1$
    & $99.1 \pm 0.0$ & $99.1 \pm 0.0$ & $99.0 \pm 0.1$
    & $\mathbf{99.1 \pm 0.0}$ & $\underline{99.1 \pm 0.0}$ & $99.0 \pm 0.0$
    & $99.0 \pm 0.0$ & $99.1 \pm 0.0$ & $99.0 \pm 0.0$ & $99.1 \pm 0.0$
    & $99.1 \pm 0.0$ \\
    \hspace{0.5em}web-Google
    & $77.6 \pm 0.1$ & $78.4 \pm 0.5$ & $79.3 \pm 0.7$ & $78.6 \pm 0.6$
    & $\underline{79.5 \pm 0.2}$ & $\mathbf{80.1 \pm 0.4}$ & $79.3 \pm 0.6$
    & $79.2 \pm 0.3$ & $79.3 \pm 0.2$ & $79.2 \pm 0.2$
    & $79.4 \pm 0.7$ & $79.4 \pm 0.1$ & $79.1 \pm 0.1$ & $78.7 \pm 0.2$
    & $77.6 \pm 0.3$ \\
    \hspace{0.5em}soc-Pokec
    & $78.0 \pm 0.7$ & $76.2 \pm 4.4$ & $82.4 \pm 0.4$ & $81.5 \pm 0.4$
    & $\mathbf{82.6 \pm 0.1}$ & $82.3 \pm 0.1$ & $82.2 \pm 0.6$
    & $82.3 \pm 0.8$ & $82.4 \pm 0.4$ & $82.4 \pm 0.2$
    & $81.3 \pm 0.5$ & $\underline{82.5 \pm 0.2}$ & $81.8 \pm 0.5$ & $77.4 \pm 0.9$
    & $80.6 \pm 1.0$ \\
    \hspace{0.5em}wiki-topcats
    & $\mathbf{62.2 \pm 1.1}$ & $58.5 \pm 4.4$ & $\underline{61.6 \pm 1.7}$ & $61.1 \pm 1.2$
    & $61.1 \pm 0.8$ & $61.5 \pm 1.4$ & $61.3 \pm 0.8$
    & $60.4 \pm 1.6$ & $61.2 \pm 1.3$ & $60.4 \pm 0.1$
    & $58.4 \pm 1.4$ & $61.5 \pm 0.8$ & $60.5 \pm 2.9$ & $54.4 \pm 1.3$
    & $59.4 \pm 1.0$ \\
    \hspace{0.5em}wiki-Talk
    & $99.0 \pm 0.1$ & $99.3 \pm 0.0$ & $99.3 \pm 0.0$ & $99.1 \pm 0.0$
    & $99.3 \pm 0.0$ & $99.3 \pm 0.0$ & $99.2 \pm 0.0$
    & $\mathbf{99.3 \pm 0.0}$ & $99.3 \pm 0.0$ & $99.2 \pm 0.1$
    & $99.2 \pm 0.0$ & $\mathbf{99.3 \pm 0.0}$ & $99.2 \pm 0.0$ & $99.3 \pm 0.0$
    & $99.2 \pm 0.0$ \\
    \hspace{0.5em}soc-LiveJournal1
    & $77.5 \pm 0.5$ & $80.1 \pm 1.0$ & $82.2 \pm 0.5$ & $80.7 \pm 0.3$
    & $\underline{82.6 \pm 0.0}$ & $82.1 \pm 0.3$ & $82.4 \pm 0.7$
    & $82.3 \pm 0.1$ & $82.3 \pm 0.2$ & $\mathbf{82.7 \pm 0.4}$
    & $81.4 \pm 0.4$ & $82.4 \pm 0.5$ & $82.1 \pm 0.1$ & $79.8 \pm 0.7$
    & $79.8 \pm 0.4$ \\
    \midrule
    \textbf{AVG}
    & $81.6 \pm 1.0$ & $83.9 \pm 1.7$ & $84.0 \pm 0.5$ & $83.1 \pm 0.8$
    & $\underline{84.7 \pm 0.4}$ & $83.9 \pm 0.8$ & $84.4 \pm 1.0$
    & $\mathbf{85.0 \pm 0.7}$ & $84.6 \pm 0.6$ & $84.6 \pm 0.5$
    & $82.6 \pm 0.8$ & $84.6 \pm 0.4$ & $84.0 \pm 0.6$ & $83.3 \pm 0.7$
    & $83.8 \pm 0.7$ \\
    \bottomrule
  \end{tabular}
  }

\end{table*}

\begin{table*}[ht]
  \centering
  \caption{Ablation: Feature Initialization (accuracy). ``random-6'' denotes a six-dimensional random feature vector 
\(\mathbf{x} \sim \mathcal{U}([0,1]^6)\).}
  \label{tab:init}
  \begin{tabular}{@{}llllllllll@{}}
    \toprule
    \textbf{Graph} & [$d^{(0)}$] & $[d^{(0)} d^{(1)}]$ & $[\dots d^{(2)}]$ & $[\dots d^{(3)}]$ & $[\dots d^{(4)}]$ & $[\dots d^{(5)}]$$^{\dagger}$ & $[\dots d^{(11)}]$ & $[d^{(0)}, Ad^{(0)}, \dots, A^5d^{(0)}]$ & \textbf{random-6} \\
    \midrule
    \multicolumn{10}{@{}l@{}}{\textbf{\textit{Undirected}}} \\
    \hspace{0.5em}p2p-Gnutella31 & $89.5 \pm 1.8$ & $\underline{90.2 \pm 0.3}$ & $\mathbf{93.2 \pm 1.2}$ & $90.1 \pm 1.0$ & $90.0 \pm 1.0$ & $89.7 \pm 0.7$ & $88.0 \pm 3.5$ & $89.6 \pm 2.0$ & $80.0 \pm 0.3$ \\
    \hspace{0.5em}com-youtube & $91.3 \pm 0.1$ & $91.0 \pm 0.4$ & $91.0 \pm 0.5$ & $91.8 \pm 0.3$ & $91.8 \pm 0.2$ & $\mathbf{92.2 \pm 0.1}$ & $91.4 \pm 0.4$ & $\underline{92.2 \pm 0.1}$ & $73.2 \pm 0.2$ \\
    \hspace{0.5em}amazon & $82.2 \pm 0.6$ & $82.2 \pm 1.5$ & $82.8 \pm 2.1$ & $82.9 \pm 1.9$ & $\mathbf{86.6 \pm 0.4}$ & $\underline{85.9 \pm 1.7}$ & $83.1 \pm 2.7$ & $85.6 \pm 1.0$ & $75.0 \pm 0.1$ \\
    \hspace{0.5em}cit-Patents & $73.0 \pm 2.2$ & $73.2 \pm 1.7$ & $72.8 \pm 0.7$ & $72.1 \pm 2.6$ & $71.2 \pm 4.1$ & $\mathbf{74.0 \pm 0.2}$ & $\underline{73.3 \pm 1.8}$ & $73.0 \pm 1.1$ & $72.5 \pm 0.1$ \\
    \hspace{0.5em}com-lj & $79.6 \pm 0.9$ & $80.1 \pm 0.7$ & $79.5 \pm 0.2$ & $79.7 \pm 0.8$ & $80.0 \pm 0.8$ & $\mathbf{80.5 \pm 0.2}$ & $79.9 \pm 0.5$ & $\underline{80.3 \pm 0.2}$ & $74.9 \pm 0.2$ \\
    \hspace{0.5em}dblp & $84.6 \pm 1.2$ & $85.5 \pm 1.8$ & $\underline{85.6 \pm 1.7}$ & $84.3 \pm 2.2$ & $\mathbf{85.6 \pm 1.2}$ & $85.0 \pm 1.3$ & $84.6 \pm 1.5$ & $83.9 \pm 1.7$ & $77.7 \pm 0.2$ \\
    \cmidrule(lr){1-10}
    \multicolumn{10}{@{}l@{}}{\textbf{\textit{Directed}}} \\
    \hspace{0.5em}soc-Epinions1 & $92.6 \pm 0.0$ & $92.5 \pm 0.1$ & $\underline{92.7 \pm 0.1}$ & $\mathbf{92.7 \pm 0.1}$ & $92.5 \pm 0.2$ & $92.6 \pm 0.1$ & $92.3 \pm 0.2$ & $92.6 \pm 0.1$ & $70.8 \pm 0.1$ \\
    \hspace{0.5em}soc-Slashdot0902 & $89.6 \pm 0.5$ & $89.7 \pm 0.5$ & $89.6 \pm 0.5$ & $89.8 \pm 0.2$ & $90.1 \pm 0.3$ & $\mathbf{90.3 \pm 0.4}$ & $89.7 \pm 0.3$ & $\underline{90.2 \pm 0.6}$ & $80.4 \pm 0.1$ \\
    \hspace{0.5em}email-EuAll & $99.0 \pm 0.0$ & $99.1 \pm 0.1$ & $99.1 \pm 0.0$ & $99.1 \pm 0.0$ & $\underline{99.1 \pm 0.0}$ & $\mathbf{99.2 \pm 0.0}$ & $99.1 \pm 0.0$ & $99.1 \pm 0.0$ & $27.4 \pm 0.0$ \\
    \hspace{0.5em}web-Google & $78.8 \pm 0.1$ & $78.8 \pm 0.3$ & $79.2 \pm 0.3$ & $\underline{79.8 \pm 0.4}$ & $79.5 \pm 0.5$ & $79.8 \pm 0.4$ & $79.4 \pm 0.3$ & $\mathbf{80.0 \pm 0.4}$ & $64.6 \pm 0.1$ \\
    \hspace{0.5em}soc-Pokec & $81.4 \pm 1.3$ & $82.5 \pm 0.5$ & $82.3 \pm 0.8$ & $81.8 \pm 0.8$ & $82.7 \pm 0.3$ & $\mathbf{83.3 \pm 0.6}$ & $82.4 \pm 1.3$ & $\underline{83.1 \pm 0.4}$ & $78.7 \pm 0.2$ \\
    \hspace{0.5em}wiki-topcats & $62.3 \pm 0.5$ & $61.9 \pm 1.5$ & $60.4 \pm 1.6$ & $60.0 \pm 2.4$ & $62.5 \pm 1.2$ & $63.1 \pm 0.9$ & $\underline{63.4 \pm 0.7}$ & $62.5 \pm 0.9$ & $\mathbf{63.5 \pm 0.1}$ \\
    \hspace{0.5em}wiki-Talk & $99.3 \pm 0.0$ & $99.3 \pm 0.0$ & $99.3 \pm 0.0$ & $99.3 \pm 0.0$ & $99.3 \pm 0.0$ & $\mathbf{99.3 \pm 0.0}$ & $99.3 \pm 0.0$ & $\underline{99.3 \pm 0.0}$ & $28.2 \pm 0.0$ \\
    \hspace{0.5em}soc-LiveJournal1 & $82.3 \pm 0.5$ & $82.3 \pm 0.5$ & $82.3 \pm 0.1$ & $82.2 \pm 0.6$ & $82.3 \pm 0.4$ & $\mathbf{82.8 \pm 0.2}$ & $82.4 \pm 0.3$ & $\underline{82.6 \pm 0.2}$ & $75.8 \pm 0.2$ \\
    \midrule
    \textbf{AVG} & $84.7 \pm 0.7$ & $84.9 \pm 0.7$ & $85.0 \pm 0.7$ & $84.7 \pm 1.0$ & $85.2 \pm 0.7$ & $\mathbf{85.5 \pm 0.5}$ & $84.9 \pm 1.0$ & $\underline{85.3 \pm 0.6}$ & $67.3 \pm 0.1$ \\
    \bottomrule
  \end{tabular}
\end{table*}
\begin{table*}[t]
  \centering
  \scriptsize
  \caption{Ablation: dropout, encoder sharing, and fusion strategy (accuracy).}
  \label{tab:dropout_fusion}
  \resizebox{\textwidth}{!}{%
  \begin{tabular}{@{}lcccccccccc@{}}
    \toprule
    & \multicolumn{6}{c}{\textbf{Dropout}}
    & \multicolumn{2}{c}{\textbf{Encoder}}
    & \multicolumn{2}{c}{\textbf{Fusion}} \\
    \cmidrule(lr){2-7} \cmidrule(lr){8-9} \cmidrule(lr){10-11}
    \textbf{Graph}
    & \textbf{0.0}
    & \textbf{0.1}
    & \textbf{0.2}
    & \textbf{0.3$^{\dagger}$}
    & \textbf{0.4}
    & \textbf{0.5}
    & \textbf{Shared$^{\dagger}$}
    & \textbf{Unshared}
    & \textbf{Add}
    & \textbf{Cat-MLP} \\
    \midrule

    \multicolumn{11}{@{}l@{}}{\textbf{\textit{Undirected}}} \\

    \hspace{0.5em}p2p-Gnutella31
    & $88.9 \pm 0.9$ & $89.0 \pm 0.8$ & $88.7 \pm 0.5$ & $89.7 \pm 0.7$ & $\mathbf{91.5 \pm 0.1}$ & $\underline{90.4 \pm 1.0}$
    & $89.7 \pm 0.7$ & $\mathbf{91.5 \pm 0.4}$ & $\underline{90.1 \pm 1.5}$ & $89.6 \pm 0.4$ \\

    \hspace{0.5em}com-youtube
    & $91.5 \pm 0.3$ & $\mathbf{92.3 \pm 0.4}$ & $92.1 \pm 0.3$ & $\underline{92.2 \pm 0.1}$ & $91.7 \pm 0.1$ & $91.3 \pm 0.4$
    & $\mathbf{92.2 \pm 0.1}$ & $91.7 \pm 0.1$ & $\underline{91.9 \pm 0.4}$ & $91.7 \pm 0.7$ \\

    \hspace{0.5em}amazon
    & $83.2 \pm 1.1$ & $\underline{85.6 \pm 0.6}$ & $84.0 \pm 1.1$ & $\mathbf{85.9 \pm 1.7}$ & $84.5 \pm 0.5$ & $83.9 \pm 2.9$
    & $\mathbf{85.9 \pm 1.7}$ & $84.8 \pm 1.0$ & $\underline{85.2 \pm 2.3}$ & $83.0 \pm 3.1$ \\

    \hspace{0.5em}cit-Patents
    & $72.8 \pm 0.8$ & $\underline{73.7 \pm 2.3}$ & $72.2 \pm 0.6$ & $\mathbf{74.0 \pm 0.2}$ & $72.9 \pm 1.5$ & $72.2 \pm 2.1$
    & $\underline{74.0 \pm 0.2}$ & $\mathbf{74.1 \pm 1.4}$ & $73.4 \pm 0.2$ & $71.9 \pm 1.2$ \\

    \hspace{0.5em}com-lj
    & $79.5 \pm 0.4$ & $\underline{80.3 \pm 1.4}$ & $80.1 \pm 1.1$ & $\mathbf{80.5 \pm 0.2}$ & $79.5 \pm 0.6$ & $79.5 \pm 0.3$
    & $\mathbf{80.5 \pm 0.2}$ & $\underline{80.5 \pm 0.5}$ & $80.5 \pm 1.0$ & $78.0 \pm 2.1$ \\

    \hspace{0.5em}dblp
    & $80.3 \pm 1.8$ & $82.8 \pm 2.0$ & $83.0 \pm 1.4$ & $85.0 \pm 1.3$ & $\underline{86.0 \pm 0.5}$ & $\mathbf{86.2 \pm 2.8}$
    & $85.0 \pm 1.3$ & $\mathbf{86.0 \pm 0.2}$ & $\underline{85.4 \pm 0.9}$ & $84.8 \pm 1.2$ \\

    \cmidrule(lr){1-11}
    \multicolumn{11}{@{}l@{}}{\textbf{\textit{Directed}}} \\

    \hspace{0.5em}soc-Epinions1
    & $92.3 \pm 0.2$ & $92.4 \pm 0.2$ & $92.4 \pm 0.2$ & $\mathbf{92.6 \pm 0.1}$ & $92.4 \pm 0.1$ & $\underline{92.5 \pm 0.1}$
    & $\underline{92.6 \pm 0.1}$ & $92.6 \pm 0.2$ & $\mathbf{92.7 \pm 0.3}$ & $92.4 \pm 0.5$ \\

    \hspace{0.5em}soc-Slashdot0902
    & $89.9 \pm 0.2$ & $90.0 \pm 0.2$ & $90.0 \pm 0.5$ & $\mathbf{90.3 \pm 0.4}$ & $\underline{90.2 \pm 0.2}$ & $89.8 \pm 0.5$
    & $\mathbf{90.3 \pm 0.4}$ & $89.5 \pm 0.3$ & $89.8 \pm 0.6$ & $\underline{89.8 \pm 0.1}$ \\

    \hspace{0.5em}email-EuAll
    & $99.0 \pm 0.1$ & $\underline{99.1 \pm 0.0}$ & $\underline{99.1 \pm 0.0}$ & $\mathbf{99.2 \pm 0.0}$ & $99.1 \pm 0.0$ & $99.1 \pm 0.0$
    & $\underline{99.2 \pm 0.0}$ & $99.1 \pm 0.0$ & $99.1 \pm 0.0$ & $\mathbf{99.2 \pm 0.0}$ \\

    \hspace{0.5em}web-Google
    & $\mathbf{80.8 \pm 0.3}$ & $\underline{80.5 \pm 0.2}$ & $80.3 \pm 0.1$ & $79.8 \pm 0.4$ & $79.5 \pm 0.1$ & $79.1 \pm 0.2$
    & $\mathbf{79.8 \pm 0.4}$ & $79.3 \pm 0.2$ & $79.2 \pm 0.3$ & $\underline{79.8 \pm 0.2}$ \\

    \hspace{0.5em}soc-Pokec
    & $83.1 \pm 0.3$ & $\mathbf{83.6 \pm 0.8}$ & $82.9 \pm 0.1$ & $\underline{83.3 \pm 0.6}$ & $82.9 \pm 0.4$ & $82.3 \pm 0.5$
    & $\mathbf{83.3 \pm 0.6}$ & $82.7 \pm 0.6$ & $\underline{82.9 \pm 0.2}$ & $82.6 \pm 0.3$ \\

    \hspace{0.5em}wiki-topcats
    & $62.0 \pm 0.9$ & $\underline{63.0 \pm 1.9}$ & $62.8 \pm 2.0$ & $\mathbf{63.1 \pm 0.9}$ & $63.0 \pm 1.0$ & $61.9 \pm 1.1$
    & $63.1 \pm 0.9$ & $\mathbf{64.1 \pm 0.4}$ & $\underline{63.2 \pm 1.2}$ & $61.6 \pm 1.5$ \\

    \hspace{0.5em}wiki-Talk
    & $99.2 \pm 0.0$ & $99.3 \pm 0.0$ & $\underline{99.3 \pm 0.0}$ & $\mathbf{99.3 \pm 0.0}$ & $99.3 \pm 0.0$ & $99.3 \pm 0.0$
    & $\mathbf{99.3 \pm 0.0}$ & $99.3 \pm 0.0$ & $99.3 \pm 0.0$ & $\underline{99.3 \pm 0.0}$ \\

    \hspace{0.5em}soc-LiveJournal1
    & $82.4 \pm 0.2$ & $\mathbf{82.9 \pm 0.6}$ & $82.7 \pm 0.6$ & $\underline{82.8 \pm 0.2}$ & $82.2 \pm 0.1$ & $82.1 \pm 0.1$
    & $\underline{82.8 \pm 0.2}$ & $82.6 \pm 0.2$ & $\mathbf{82.9 \pm 0.4}$ & $81.5 \pm 1.3$ \\

    \midrule
    \textbf{AVG}
    & $84.6 \pm 0.5$ & $85.3 \pm 0.8$ & $85.0 \pm 0.6$ & $\mathbf{85.5 \pm 0.5}$ & $\underline{85.3 \pm 0.4}$ & $85.0 \pm 0.9$
    & $\underline{85.5 \pm 0.5}$ & $\mathbf{85.6 \pm 0.4}$ & $85.4 \pm 0.7$ & $84.6 \pm 0.9$ \\

    \bottomrule
  \end{tabular}%
  }
\end{table*}
\begin{table}[t]
\centering
\begin{minipage}[t]{0.53\textwidth}
  \centering
  \caption{Ablation: Depth $\times$ Width (accuracy).}
  \label{tab:depth_width}
  \small
  \begin{tabular}{@{}lllll@{}}
    \toprule
    \textbf{$L$ \textbackslash{} $n_{hid}$} & \textbf{8} & \textbf{12}$^{\dagger}$ & \textbf{16} & \textbf{32} \\
    \midrule
    \textbf{0} & $81.5 \pm 1.0$ & $81.8 \pm 0.5$ & $81.7 \pm 0.8$ & $81.8 \pm 0.6$ \\
    \textbf{1} & $84.9 \pm 0.7$ & $84.1 \pm 1.9$ & $85.0 \pm 0.9$ & $85.1 \pm 0.9$ \\
    \textbf{2}$^{\dagger}$ & $\mathbf{85.5 \pm 0.5}$ & $\underline{85.5 \pm 0.5}$ & $85.2 \pm 0.9$ & $85.4 \pm 0.8$ \\
    \textbf{4} & $84.7 \pm 0.6$ & $84.8 \pm 0.8$ & $85.5 \pm 0.6$ & $85.5 \pm 0.8$ \\
    \textbf{6} & $84.6 \pm 0.6$ & $84.8 \pm 0.6$ & $84.7 \pm 0.9$ & $85.4 \pm 0.9$ \\
    \bottomrule
  \end{tabular}
\end{minipage}
\end{table}

\begin{acks}
Experiments presented in this paper were carried out using the Grid'5000 testbed, supported by a scientific interest group hosted by Inria and including CNRS, RENATER, several universities, and other organizations (see \url{https://www.grid5000.fr}).

The research was supported by the French government National Research Agency (ANR) through UCA JEDI (ANR-15-IDEX-01) and EUR DS4H (ANR-17-EURE-004); by the France 2030 program through the PEPR CLOUD PC CARECLOUD under Grant Agreement No. ANR-23-PECL-0003 and through 3IA Côte d’Azur Investments under project reference ANR-23-IACL-000; and by the European Union through the dAIEDGE project under Grant Agreement No. 101120726 and the EURO-3C project under Grant Agreement No. 101297599.

This work was funded by the Deutsche Forschungsgemeinschaft (DFG, German Research Foundation) as part of the CRC 1639 NuMeriQS – project no. 511713970.

\end{acks}

\appendix

\section{Ablation Study}\label{sec:ablation}
To assess the impact of the main design choices of BRAVA-GNN, we
perform ablation studies on the initial node features, the synthetic
training distribution, the number of message-passing layers, the
hidden dimension, the fusion mechanism, and the dropout rate.
All results are averaged over three independent runs. The final
configuration is selected by balancing accuracy and parameter
efficiency; selected hyperparameters are marked with a \(\dagger\).

\paragraph{Training distribution.}

Since our goal is to train a model that generalizes to both directed and undirected graphs, we construct the training set using synthetic graphs from both regimes.
We start from directed Bollobás scale-free graphs (\textbf{BO}) \cite{bollobas2003directed}, following \cite{Maurya_2019}, and consider several additional graph families: symmetrized Bollobás scale-free graphs (\textbf{SBO}) described in Section~\ref{sec:trainingset}, Barabási--Albert graphs (\textbf{BA}) \cite{barabasi1999emergence}, hyperbolic random graphs (\textbf{HY}) as described in Section~\ref{sec:hyperbolic}, and directed variants of the hyperbolic model.
For the latter, we test both a uniformly directed version, where each undirected hyperbolic edge is assigned a random orientation (\textbf{UDHY}), and the directed \(m\)-nearest-neighbor (\textbf{\textit{m}NN}) construction proposed in \cite{kasyanov2023nearest}.
In the \(m\)NN variant, each node creates outgoing edges toward its \(m\) closest nodes in the hyperbolic space.
We sample \(m\) uniformly from $\{12,16,20\}$, since smaller values often produce disconnected graphs.

We evaluate multiple training mixtures, reported in Table~\ref{tab:train_distribution}.
We first train \textbf{BO, SBO, HY, UDHY} independently.
We then group the latter training sets according to edge orientation, obtaining a directed-only mix (\textbf{Dir-only}: \textbf{BO, UDHY}), an undirected-only mix (\textbf{Undir-only}: \textbf{SBO, HY}), and a combined mix containing all four models (\textbf{All-4}).
Starting from \textbf{All-4}, we also evaluate leave-one-out variants, each obtained by removing one graph model; for example, \textbf{-HY} denotes the mixture containing all families except \textbf{HY}, and corresponds to the final configuration selected for BRAVA-GNN ($\dagger$).
Finally, we assess alternative generation choices: \textbf{mNN-swap} replaces \textbf{UDHY} with the \(m\)NN hyperbolic variant, \textbf{BO+SBO+mNN} combines Bollobás scale-free graphs with the \(m\)NN variant, and \textbf{BA-swap} replaces \textbf{SBO} with graphs generated by the Barabási--Albert preferential attachment model.

\paragraph{Feature initialization.}
Table~\ref{tab:init} shows that degree-mass features are crucial for BRAVA-GNN.
Using only the degree is already competitive, but adding higher-order
degree masses improves ranking accuracy. The best performance is
obtained by using degree masses up to order five, while using higher
orders gives no further benefit. As a control, we also consider
\emph{random-6}, where each node \(v\) is initialized as
\[
\mathbf{x}^{\mathrm{rand}}_v =
[u_{v,1}, \ldots, u_{v,6}],
\qquad
u_{v,i} \overset{\mathrm{i.i.d.}}{\sim} \mathcal{U}(0,1).
\]
The lower performance of this baseline confirms that the gain comes
from the structural information encoded by degree masses, not only
from the GNN architecture.

\paragraph{Architecture.}
The remaining ablations show that a small architecture is
sufficient. Increasing the number of layers or the hidden dimension
provides only marginal gains compared to the increase in model
size. Accuracy is flat across widths at \(L=2\); we use \(n_{hid}=12\), which gives 1,297 parameters. Shared
incoming and outgoing encoders further reduce the number of
parameters with negligible loss, while multiplicative fusion performs
better than concatenation, supporting its use to combine incoming
and outgoing flow information.

Overall, the ablation results support the three main choices of
BRAVA-GNN: multi-hop degree-mass features, mixed synthetic
training graphs, and a compact directional architecture.

\section{Additional information on datasets}
Table~\ref{tab:datasets} lists the number of nodes and edges for all the graphs in the test set. 

\begin{table}[!t]
  \centering
  \caption{Node and edge counts for the 14 real-world test graphs.}
  \label{tab:datasets}
  \small
  \begin{tabular}{@{}lrr@{}}
    \toprule
    \textbf{Graph} & \textbf{\#Nodes} & \textbf{\#Edges} \\
    \midrule
    \multicolumn{3}{@{}l}{\textit{Undirected}} \\
    \hspace{0.5em}p2p-Gnutella31 & 62,586 & 147,892 \\
    \hspace{0.5em}com-youtube & 1,134,890 & 2,987,624 \\
    \hspace{0.5em}amazon & 2,146,057 & 5,743,146 \\
    \hspace{0.5em}cit-Patents & 3,764,117 & 16,511,741 \\
    \hspace{0.5em}com-lj & 3,997,962 & 34,681,189 \\
    \hspace{0.5em}dblp & 4,000,148 & 8,649,011 \\
    \cmidrule(lr){1-3}
    \multicolumn{3}{@{}l}{\textit{Directed}} \\
    \hspace{0.5em}soc-Epinions1 & 75,879 & 508,837 \\
    \hspace{0.5em}soc-Slashdot0902 & 82,168 & 948,464 \\
    \hspace{0.5em}email-EuAll & 265,214 & 420,045 \\
    \hspace{0.5em}web-Google & 875,713 & 5,105,039 \\
    \hspace{0.5em}soc-Pokec & 1,632,803 & 30,622,564 \\
    \hspace{0.5em}wiki-topcats & 1,791,489 & 28,511,807 \\
    \hspace{0.5em}wiki-Talk & 2,394,385 & 5,021,410 \\
    \hspace{0.5em}soc-LiveJournal1 & 4,847,571 & 68,993,773 \\
    \bottomrule
  \end{tabular}
\end{table}

\section{GenAI Usage Disclosure}
Generative AI tools were used as writing and coding assistants.
Specifically, they supported language refinement, code checking, and debugging.
All scientific contributions, experiments, results, and conclusions were independently verified by the authors, who take full responsibility for the final content of the manuscript.

\bibliographystyle{ACM-Reference-Format}
\balance
\bibliography{CIKM_bib}

@inproceedings{paszke2019pytorch,
  author    = {Paszke, Adam and Gross, Sam and Massa, Francisco and Lerer, Adam and
               Bradbury, James and Chanan, Gregory and Killeen, Trevor and Lin, Zeming and
               Gimelshein, Natalia and Antiga, Luca and Desmaison, Alban and
               K{\"o}pf, Andreas and Yang, Edward and DeVito, Zachary and
               Raison, Martin and Tejani, Alykhan and Chilamkurthy, Sasank and
               Steiner, Benoit and Fang, Lu and Bai, Junjie and Chintala, Soumith},
  title     = {{PyTorch}: An Imperative Style, High-Performance Deep Learning Library},
  booktitle = {Advances in Neural Information Processing Systems 32 (NeurIPS 2019)},
  publisher = {Curran Associates, Inc.},
  address   = {Red Hook, NY, USA},
  pages     = {8024--8035},
  year      = {2019}
}

@inproceedings{zhangEfficientApproximationAlgorithms2023,
  title = {Efficient {{Approximation Algorithms}} for {{Spanning Centrality}}},
  booktitle = {Proceedings of the 29th {{ACM SIGKDD Conference}} on {{Knowledge Discovery}} and {{Data Mining}}},
  author = {Zhang, Shiqi and Yang, Renchi and Tang, Jing and Xiao, Xiaokui and Tang, Bo},
  year = 2023,
  month = aug,
  series = {{{KDD}} '23},
  pages = {3386--3395},
  publisher = {Association for Computing Machinery},
  address = {New York, NY, USA},
  doi = {10.1145/3580305.3599323},
  isbn = {979-8-4007-0103-0}
}

@InProceedings{Hagberg_2008,
author =       {Aric A. Hagberg and Daniel A. Schult and Pieter J. Swart},
title =        {Exploring Network Structure, Dynamics, and Function using NetworkX},
booktitle =   {Proceedings of the 7th Python in Science Conference},
pages =     {11 - 15},
publisher = {SciPy},
address = {Pasadena, CA USA},
year =      {2008},
doi = {10.25080/tcwv9851},
editor =    {Ga\"el Varoquaux and Travis Vaught and Jarrod Millman},
}

@article{Staudt_2015,
  author  = {Staudt, Christian L. and Sazonovs, Aleksejs and Meyerhenke, Henning},
  title   = {{NetworKit}: A Tool Suite for Large-Scale Complex Network Analysis},
  journal = {Network Science},
  volume  = {4},
  number  = {4},
  pages   = {508--530},
  year    = {2016},
  doi     = {10.1017/nws.2016.20}
}

@article{Maurya_2021,
author = {Maurya, Sunil Kumar and Liu, Xin and Murata, Tsuyoshi},
title = {Graph Neural Networks for Fast Node Ranking Approximation},
year = {2021},
issue_date = {October 2021},
publisher = {Association for Computing Machinery},
address = {New York, NY, USA},
volume = {15},
number = {5},
issn = {1556-4681},
url = {https://doi.org/10.1145/3446217},
doi = {10.1145/3446217},
journal = {ACM Trans. Knowl. Discov. Data},
month = may,
articleno = {78},
numpages = {32}
}

@article{Li_2014,
  author  = {Li, Cong and Li, Qian and Van Mieghem, Piet and Stanley, H. Eugene and Wang, Huijuan},
  title   = {Correlation between Centrality Metrics and Their Application to the Opinion Model},
  journal = {The European Physical Journal B},
  volume  = {88},
  number  = {3},
  pages   = {1--13},
  year    = {2015},
  doi     = {10.1140/epjb/e2015-50671-y}
}

@article{Krioukov_2010,
  title = {Hyperbolic geometry of complex networks},
  author = {Krioukov, Dmitri and Papadopoulos, Fragkiskos and Kitsak, Maksim and Vahdat, Amin and Bogu\~n\'a, Mari\'an},
  journal = {Phys. Rev. E},
  volume = {82},
  issue = {3},
  pages = {036106},
  numpages = {18},
  year = {2010},
  month = {Sep},
  publisher = {American Physical Society},
  doi = {10.1103/PhysRevE.82.036106},
  url = {https://link.aps.org/doi/10.1103/PhysRevE.82.036106}
}

@article{Freeman_1977,
  title={A Set of Measures of Centrality Based on Betweenness},
  volume={40}, 
  ISSN={00380431}, 
  url={https://www.jstor.org/stable/3033543?origin=crossref}, DOI={10.2307/3033543}, 
  number={1}, 
  journal={Sociometry}, 
  author={Freeman, Linton C.}, 
  year={1977}, 
  month=mar, 
  pages={35} }

@article{Brandes01062001,
author = {Ulrik Brandes},
title = {A faster algorithm for betweenness centrality},
journal = {The Journal of Mathematical Sociology},
volume = {25},
number = {2},
pages = {163--177},
year = {2001},
publisher = {Routledge},
doi = {10.1080/0022250X.2001.9990249},
URL = {https://doi.org/10.1080/0022250X.2001.9990249
},
}

@inproceedings{Heeg_2024,
author = {Heeg, Franziska and Scholtes, Ingo},
title = {Using time-aware graph neural networks to predict temporal centralities in dynamic graphs},
year = {2024},
isbn = {9798331314385},
publisher = {Curran Associates Inc.},
address = {Red Hook, NY, USA},
booktitle = {Proceedings of the 38th International Conference on Neural Information Processing Systems},
articleno = {949},
numpages = {30},
location = {Vancouver, BC, Canada},
series = {NIPS '24}
}

@inproceedings{Geisberger2008,
author = {Robert Geisberger and Peter Sanders and Dominik Schultes},
title = {Better Approximation of Betweenness Centrality},
booktitle = {2008 Proceedings of the Workshop on Algorithm Engineering and Experiments (ALENEX)},
year = {2008},
pages = {90--100},
publisher = {Society for Industrial and Applied Mathematics},
address = {Philadelphia, PA, USA},
doi = {10.1137/1.9781611972887.9},
URL = {https://epubs.siam.org/doi/abs/10.1137/1.9781611972887.9},
}

@article{brandes2007centrality,
  title={Centrality estimation in large networks},
  author={Brandes, Ulrik and Pich, Christian},
  journal={International Journal of Bifurcation and Chaos},
  volume={17},
  number={07},
  pages={2303--2318},
  year={2007},
  publisher={World Scientific}
}

@inproceedings{Riondato_2014,
author = {Riondato, Matteo and Kornaropoulos, Evgenios M.},
title = {Fast approximation of betweenness centrality through sampling},
year = {2014},
isbn = {9781450323512},
publisher = {Association for Computing Machinery},
address = {New York, NY, USA},
url = {https://doi.org/10.1145/2556195.2556224},
doi = {10.1145/2556195.2556224},
booktitle = {Proceedings of the 7th ACM International Conference on Web Search and Data Mining},
pages = {413–422},
numpages = {10},
location = {New York, New York, USA},
series = {WSDM '14}
}

@article{Borassi_2019,
author = {Borassi, Michele and Natale, Emanuele},
title = {KADABRA is an ADaptive Algorithm for Betweenness via Random Approximation},
year = {2019},
issue_date = {2019},
publisher = {Association for Computing Machinery},
address = {New York, NY, USA},
volume = {24},
issn = {1084-6654},
url = {https://doi.org/10.1145/3284359},
doi = {10.1145/3284359},
journal = {ACM J. Exp. Algorithmics},
month = feb,
articleno = {1.2},
numpages = {35}
}

@article{mirakyan_2021,
  author  = {Mirakyan, Martin},
  title   = {{ABCDE}: Approximating Betweenness-Centrality Ranking with Progressive-{DropEdge}},
  journal = {PeerJ Computer Science},
  volume  = {7},
  pages   = {e699},
  year    = {2021},
  doi     = {10.7717/peerj-cs.699}
}

@inproceedings{fan2019learning,
  title={Learning to Identify High Betweenness Centrality Nodes from Scratch: A Novel Graph Neural Network Approach},
  author={Fan, Changjun and Zeng, Li and Ding, Yuhui and Chen, Muhao and Sun, Yizhou and Liu, Zhong},
  booktitle={Proceedings of the 28th ACM International Conference on Information and Knowledge Management},
  pages={559--568},
  year={2019},
  publisher={Association for Computing Machinery},
  address={New York, NY, USA},
  location={Beijing, China},
  series={CIKM '19},
  doi={10.1145/3357384.3357979}
}

@inproceedings{Maurya_2019,
author = {Maurya, Sunil Kumar and Liu, Xin and Murata, Tsuyoshi}, 
title = {Fast Approximations of Betweenness Centrality with Graph Neural Networks},
year = {2019},
isbn = {9781450369763}, 
publisher = {Association for Computing Machinery}, address = {New York, NY, USA}, 
url = {https://doi.org/10.1145/3357384.3358080}, doi = {10.1145/3357384.3358080},
booktitle = {Proceedings of the 28th ACM International Conference on Information and Knowledge Management},
  pages = {2149–2152},
  numpages = {4}, 
    location = {Beijing, China}, series = {CIKM '19} }

@INPROCEEDINGS{Hua_2016,
  author={Hua, Qiang-Sheng and Fan, Haoqiang and Ai, Ming and Qian, Lixiang and Li, Yangyang and Shi, Xuanhua and Jin, Hai},
  booktitle={2016 IEEE 36th International Conference on Distributed Computing Systems (ICDCS)}, 
  title={Nearly Optimal Distributed Algorithm for Computing Betweenness Centrality}, 
  year={2016},
  volume={},
  number={},
  pages={271-280},
  publisher={IEEE},
  address={Piscataway, NJ, USA},
  doi={10.1109/ICDCS.2016.89}}

@inproceedings{Hoang_2019, 
author = {Hoang, Loc and Pontecorvi, Matteo and Dathathri, Roshan and Gill, Gurbinder and You, Bozhi and Pingali, Keshav and Ramachandran, Vijaya}, 
title = {A round-efficient distributed betweenness centrality algorithm},
year = {2019},
isbn = {9781450362252},
publisher = {Association for Computing Machinery}, 
address = {New York, NY, USA}, 
url = {https://doi.org/10.1145/3293883.3295729},
doi = {10.1145/3293883.3295729}, 
booktitle = {Proceedings of the 24th Symposium on Principles and Practice of Parallel Programming},
pages = {272–286}, 
numpages = {15}, 
location = {Washington, District of Columbia}, 
series = {PPoPP '19} }

@inproceedings{Chen_2009,
 author = {Chen, Wei and Liu, Tie-yan and Lan, Yanyan and Ma, Zhi-ming and Li, Hang},
 booktitle = {Advances in Neural Information Processing Systems},
 editor = {Y. Bengio and D. Schuurmans and J. Lafferty and C. Williams and A. Culotta},
 pages = {},
 publisher = {Curran Associates, Inc.},
 address = {Red Hook, NY, USA},
 title = {Ranking Measures and Loss Functions in Learning to Rank},
 url = {https://proceedings.neurips.cc/paper_files/paper/2009/file/2f55707d4193dc27118a0f19a1985716-Paper.pdf},
 volume = {22},
 year = {2009}
}

@article{10.1145/3208351,
author = {Riondato, Matteo and Upfal, Eli},
title = {ABRA: Approximating Betweenness Centrality in Static and Dynamic Graphs with Rademacher Averages},
year = {2018},
issue_date = {October 2018},
publisher = {Association for Computing Machinery},
address = {New York, NY, USA},
volume = {12},
number = {5},
issn = {1556-4681},
url = {https://doi.org/10.1145/3208351},
doi = {10.1145/3208351},
journal = {ACM Trans. Knowl. Discov. Data},
month = jul,
articleno = {61},
numpages = {38}
}

@inproceedings{10.1145/2939672.2939754,
author = {Grover, Aditya and Leskovec, Jure},
title = {node2vec: Scalable Feature Learning for Networks},
year = {2016},
isbn = {9781450342322},
publisher = {Association for Computing Machinery},
address = {New York, NY, USA},
url = {https://doi.org/10.1145/2939672.2939754},
doi = {10.1145/2939672.2939754},
booktitle = {Proceedings of the 22nd ACM SIGKDD International Conference on Knowledge Discovery and Data Mining},
pages = {855–864},
numpages = {10},
location = {San Francisco, California, USA},
series = {KDD '16}
}

@article{BORASSI201651,
title = {Into the Square: On the Complexity of Some Quadratic-time Solvable Problems},
journal = {Electronic Notes in Theoretical Computer Science},
volume = {322},
pages = {51-67},
year = {2016},
note = {Proceedings of ICTCS 2015, the 16th Italian Conference on Theoretical Computer Science},
issn = {1571-0661},
doi = {10.1016/j.entcs.2016.03.005},
url = {https://www.sciencedirect.com/science/article/pii/S1571066116300135},
author = {Michele Borassi and Pierluigi Crescenzi and Michel Habib}
}

@article{IMPAGLIAZZO2001512,
title = {Which Problems Have Strongly Exponential Complexity?},
journal = {Journal of Computer and System Sciences},
volume = {63},
number = {4},
pages = {512-530},
year = {2001},
issn = {0022-0000},
doi = {10.1006/jcss.2001.1774},
url = {https://www.sciencedirect.com/science/article/pii/S002200000191774X},
author = {Russell Impagliazzo and Ramamohan Paturi and Francis Zane}
}

@inproceedings{Mahmoody_2016,
author = {Mahmoody, Ahmad and Tsourakakis, Charalampos E. and Upfal, Eli},
title = {Scalable Betweenness Centrality Maximization via Sampling},
year = {2016},
isbn = {9781450342322},
publisher = {Association for Computing Machinery},
address = {New York, NY, USA},
url = {https://doi.org/10.1145/2939672.2939869},
doi = {10.1145/2939672.2939869},
booktitle = {Proceedings of the 22nd ACM SIGKDD International Conference on Knowledge Discovery and Data Mining},
pages = {1765–1773},
numpages = {9},
location = {San Francisco, California, USA},
series = {KDD '16}
}

@inproceedings{Yoshida_2014,
author = {Yoshida, Yuichi},
title = {Almost linear-time algorithms for adaptive betweenness centrality using hypergraph sketches},
year = {2014},
isbn = {9781450329569},
publisher = {Association for Computing Machinery},
address = {New York, NY, USA},
url = {https://doi.org/10.1145/2623330.2623626},
doi = {10.1145/2623330.2623626},
booktitle = {Proceedings of the 20th ACM SIGKDD International Conference on Knowledge Discovery and Data Mining},
pages = {1416–1425},
numpages = {10},
location = {New York, New York, USA},
series = {KDD '14}
}

@inproceedings{Brunelli_2024,
author = {Brunelli, Filippo and Crescenzi, Pierluigi and Viennot, Laurent},
title = {Making Temporal Betweenness Computation Faster and Restless},
year = {2024},
isbn = {9798400704901},
publisher = {Association for Computing Machinery},
address = {New York, NY, USA},
url = {https://doi.org/10.1145/3637528.3671825},
doi = {10.1145/3637528.3671825},
booktitle = {Proceedings of the 30th ACM SIGKDD Conference on Knowledge Discovery and Data Mining},
pages = {163–174},
numpages = {12},
location = {Barcelona, Spain},
series = {KDD '24}
}

@InProceedings{VonLooz_2016,
author="von Looz, Moritz
and Meyerhenke, Henning",
editor="M{\"a}kinen, Veli
and Puglisi, Simon J.
and Salmela, Leena",
title="Querying Probabilistic Neighborhoods in Spatial Data Sets Efficiently",
booktitle="Combinatorial Algorithms",
year="2016",
publisher="Springer International Publishing",
address="Cham",
pages="449--460",
isbn="978-3-319-44543-4"
}

@InProceedings{Gugelmann_2012,
author="Gugelmann, Luca
and Panagiotou, Konstantinos
and Peter, Ueli",
editor="Czumaj, Artur
and Mehlhorn, Kurt
and Pitts, Andrew
and Wattenhofer, Roger",
title="Random Hyperbolic Graphs: Degree Sequence and Clustering",
booktitle="Automata, Languages, and Programming",
year="2012",
publisher="Springer Berlin Heidelberg",
address="Berlin, Heidelberg",
pages="573--585",
isbn="978-3-642-31585-5"
}

@article{MULLER_2019,
 ISSN = {00018678},
 URL = {http://www.jstor.org/stable/45277962},
 author = {Tobias Müller and Merlijn Staps},
 journal = {Advances in Applied Probability},
 number = {2},
 pages = {358--377},
 publisher = {Applied Probability Trust},
 title = {The diameter of KPKVB random graphs},
 urldate = {2026-02-04},
 volume = {51},
 year = {2019}
}

@article{Friedrich_2018,
author = {Friedrich, Tobias and Krohmer, Anton},
title = {On the Diameter of Hyperbolic Random Graphs},
journal = {SIAM Journal on Discrete Mathematics},
volume = {32},
number = {2},
pages = {1314-1334},
year = {2018},
doi = {10.1137/17M1123961},
}

@inproceedings{Network_Data_Repo,
     title={The Network Data Repository with Interactive Graph Analytics and Visualization},
     author={Ryan A. Rossi and Nesreen K. Ahmed},
     booktitle={Proceedings of the Twenty-Ninth AAAI Conference on Artificial Intelligence},
     pages={4292--4293},
     publisher={AAAI Press},
     address={Palo Alto, CA, USA},
     doi={10.1609/aaai.v29i1.9277},
     url={https://networkrepository.com},
     year={2015}
}

@inproceedings{Solomonik_2017, 
author = {Solomonik, Edgar and Besta, Maciej and Vella, Flavio and Hoefler, Torsten}, 
title = {Scaling betweenness centrality using communication-efficient sparse matrix multiplication},
year = {2017},
isbn = {9781450351140},
publisher = {Association for Computing Machinery},
address = {New York, NY, USA}, 
url = {https://doi.org/10.1145/3126908.3126971}, 
doi = {10.1145/3126908.3126971}, 
booktitle = {Proceedings of the International Conference for High Performance Computing, Networking, Storage and Analysis}, 
numpages = {14},
location = {Denver, Colorado}, 
series = {SC '17} }

@inproceedings{Bernaschi_2016,
author = {Bernaschi, Massimo and Carbone, Giancarlo and Vella, Flavio},
title = {Scalable betweenness centrality on multi-GPU systems},
year = {2016},
isbn = {9781450341288},
publisher = {Association for Computing Machinery},
address = {New York, NY, USA},
url = {https://doi.org/10.1145/2903150.2903153},
doi = {10.1145/2903150.2903153},
booktitle = {Proceedings of the ACM International Conference on Computing Frontiers},
pages = {29–36},
numpages = {8},
location = {Como, Italy},
series = {CF '16}
}

@article{Vella_2018,
author = {Vella, Flavio and Bernaschi, Massimo and Carbone, Giancarlo},
title = {Dynamic Merging of Frontiers for Accelerating the Evaluation of Betweenness Centrality},
year = {2018},
issue_date = {2018},
publisher = {Association for Computing Machinery},
address = {New York, NY, USA},
volume = {23},
issn = {1084-6654},
url = {https://doi.org/10.1145/3182656},
doi = {10.1145/3182656},

journal = {ACM J. Exp. Algorithmics},
month = mar,
articleno = {1.4},
numpages = {19}
}

@misc{snapnets,
    author       = {Jure Leskovec and Andrej Krevl},
    title        = {{SNAP Datasets}: {Stanford} Large Network Dataset Collection},
    howpublished = {\url{http://snap.stanford.edu/data}},
    month        = jun,
    year         = 2014
}

@inproceedings{AlGhamdi_2017,
author = {AlGhamdi, Ziyad and Jamour, Fuad and Skiadopoulos, Spiros and Kalnis, Panos},
title = {A Benchmark for Betweenness Centrality Approximation Algorithms on Large Graphs},
year = {2017},
isbn = {9781450352826},
publisher = {Association for Computing Machinery},
address = {New York, NY, USA},
url = {https://doi.org/10.1145/3085504.3085510},
doi = {10.1145/3085504.3085510},
booktitle = {Proceedings of the 29th International Conference on Scientific and Statistical Database Management},
articleno = {6},
numpages = {12},
location = {Chicago, IL, USA},
series = {SSDBM '17}
}

@article{lee_2021,
  author  = {Lee, Jongshin and Lee, Yongsun and Oh, Soo Min and Kahng, B.},
  title   = {Betweenness Centrality of Teams in Social Networks},
  journal = {Chaos: An Interdisciplinary Journal of Nonlinear Science},
  volume  = {31},
  number  = {6},
  pages   = {061108},
  year    = {2021},
  doi     = {10.1063/5.0056683}
}

@article{Behera_2019,
  author  = {Behera, Ranjan Kumar and Rath, Santanu Kumar and Misra, Sanjay and
             Dama{\v{s}}evi{\v{c}}ius, Robertas and Maskeli{\=u}nas, Rytis},
  title   = {Distributed Centrality Analysis of Social Network Data Using {MapReduce}},
  journal = {Algorithms},
  volume  = {12},
  number  = {8},
  pages   = {161},
  year    = {2019},
  doi     = {10.3390/a12080161}
}

@article{kasyanov2023nearest,
  author  = {Kasyanov, I. A. and van der Hoorn, Pim and Krioukov, Dmitri and Tamm, M. V.},
  title   = {Nearest-Neighbor Directed Random Hyperbolic Graphs},
  journal = {Physical Review E},
  volume  = {108},
  number  = {5},
  pages   = {054310},
  year    = {2023},
  doi     = {10.1103/PhysRevE.108.054310}
}

@article{pellegrina_2023,
author = {Pellegrina, Leonardo and Vandin, Fabio},
title = {
SILVAN: Estimating Betweenness Centralities with Progressive Sampling and Non-uniform Rademacher Bounds},
year = {2023},
issue_date = {April 2024},
publisher = {Association for Computing Machinery},
address = {New York, NY, USA},
volume = {18},
number = {3},
issn = {1556-4681},
url = {https://doi.org/10.1145/3628601},
doi = {10.1145/3628601},
journal = {ACM Trans. Knowl. Discov. Data},
month = dec,
articleno = {52},
numpages = {55}
}

@article{Cousins_2023_Bavarian,
author = {Cousins, Cyrus and Wohlgemuth, Chloe and Riondato, Matteo},
title = {Bavarian: Betweenness Centrality Approximation with Variance-Aware Rademacher Averages},
year = {2023},
publisher = {Association for Computing Machinery},
address = {New York, NY, USA},
volume = {17},
number = {6},
issn = {1556-4681},
url = {https://doi.org/10.1145/3577021},
doi = {10.1145/3577021},
journal = {ACM Trans. Knowl. Discov. Data},
month = mar,
articleno = {78},
numpages = {47}
}

@article{barabasi1999emergence,
  author  = {Barab{\'a}si, Albert-L{\'a}szl{\'o} and Albert, R{\'e}ka},
  title   = {Emergence of Scaling in Random Networks},
  journal = {Science},
  volume  = {286},
  number  = {5439},
  pages   = {509--512},
  year    = {1999},
  doi     = {10.1126/science.286.5439.509}
}

@inproceedings{bollobas2003directed,
  title     = {Directed Scale-Free Graphs},
  author    = {Bollob{\'a}s, B{\'e}la and Borgs, Christian and Chayes, Jennifer and Riordan, Oliver},
  booktitle = {Proceedings of the Fourteenth Annual ACM-SIAM Symposium on Discrete Algorithms},
  series    = {SODA '03},
  pages     = {132--139},
  year      = {2003},
  publisher = {Society for Industrial and Applied Mathematics},
  address   = {Philadelphia, PA, USA}
}

\end{document}